%% file: main.tex
\documentclass{article} %
\input{preamble}

\DeclareRobustCommand{\alligatoremoji}{%
    \raisebox{-0.15em}{\includegraphics[height=1.1em]{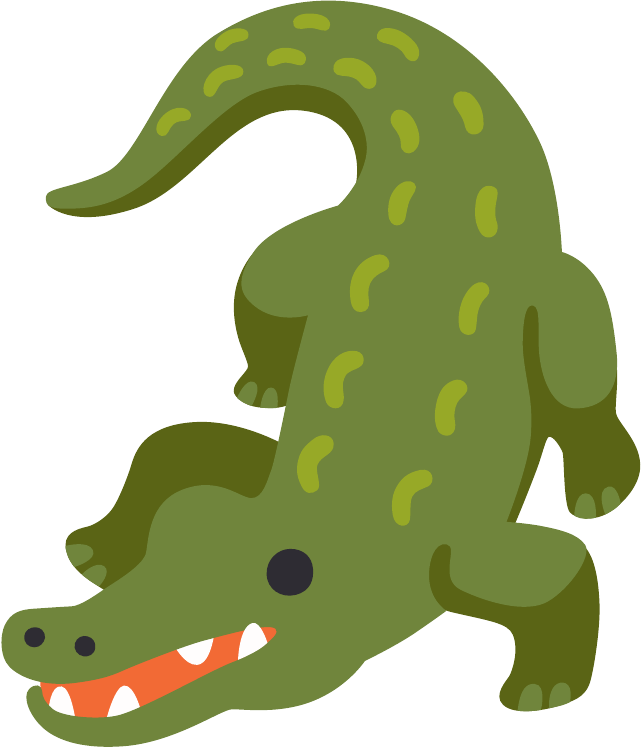}}%
}
\title{\alligatoremoji\ \modelname{}: Generative and Agentic 3D Object \mbox{Reconstruction} From Casual Images}
\hypersetup{pdftitle={\modelname{}: Generative and Agentic 3D Object Reconstruction From Casual Images}}
\author{Qirui Wu$^{1,2}$\quad Stan Birchfield$^{1}$\quad Hesam Rabeti$^{1}$\quad Angel X.~Chang$^{2}$\quad Bowen Wen$^{1}$\\
$^{1}$NVIDIA\quad $^{2}$Simon Fraser University}

\newif\ifarxiv
\arxivtrue
\ifarxiv
\iclrfinalcopy
\hypersetup{pdfauthor={Qirui Wu, Stan Birchfield, Hesam Rabeti, Angel X. Chang, Bowen Wen}}
\fi
\begin{document}

\maketitle
\ifarxiv\lhead{Preprint}\fi

\input{figures/teaser}

\input{sections/abstract}

\input{sections/introduction}

\input{sections/related_work}

\input{sections/approach}

\input{sections/experiments}

\input{sections/conclusion}

\FloatBarrier

\bibliography{ref}
\bibliographystyle{iclr2027_conference}

\clearpage
\appendix
\input{sections/supplement}

\end{document}

%% file: preamble.tex
\usepackage[T1]{fontenc}
\usepackage{textcomp}
\usepackage{xcolor}
\usepackage{iclr2027_conference,times}

\usepackage{amsmath}
\usepackage{amssymb}
\usepackage{mathtools}
\usepackage{graphicx}
\usepackage{tikz}
\usepackage{import}
\usepackage{array}
\usepackage{tabularx}
\usepackage{booktabs}
\usepackage{bigstrut}
\usepackage{colortbl}
\usepackage{hhline}
\usepackage{makecell}
\usepackage{multirow}
\usepackage{wrapfig,lipsum}
\usepackage{rotating}
\usepackage{placeins}
\usepackage{stfloats}
\usepackage{gensymb}
\usepackage{pifont} %
\usepackage[nointegrals]{wasysym}
\SetSymbolFont{wasy}{bold}{U}{wasy}{m}{n} %
\usepackage{yfonts}
\usepackage{soul}
\usepackage{appendix}
\usepackage{titletoc} %
\usepackage{setspace}
\usepackage{atbegshi,picture}
\usepackage[font={footnotesize}]{caption} %
\usepackage{listings} %

\lstdefinestyle{prompt}{basicstyle=\ttfamily\scriptsize, breaklines=true, breakatwhitespace=true, columns=fullflexible, keepspaces=true, frame=single, framesep=4pt, rulecolor=\color{black!30}, xleftmargin=4pt, xrightmargin=4pt, aboveskip=8pt, belowskip=8pt}

\newcommand{\modelname}{\textup{GATOR}}
\newcommand{\addsbrecall}{\ensuremath{\text{ADD-SB@0.1}}}

\newcommand{\boldparagraphstart}[1]{\vspace{1pt}\noindent \textbf{#1}}
\definecolor{nvidiagreen}{HTML}{76B900}
\newcommand{\bestresult}[1]{\cellcolor{nvidiagreen!80}\textbf{#1}}
\newcommand{\secondresult}[1]{\cellcolor{nvidiagreen!25}#1}
\definecolor{darkgreen}{RGB}{0,127,0}
\definecolor{darkred}{RGB}{200,0,0}

\input{math_commands.tex}

\usepackage[accsupp]{axessibility} %
\usepackage{hyperref}
\usepackage{url}

%% file: math_commands.tex
\usepackage{amsmath,amsfonts,bm}

\def\eqref#1{equation~\ref{#1}}

\def\1{\bm{1}}

\DeclareMathAlphabet{\mathsfit}{\encodingdefault}{\sfdefault}{m}{sl}
\SetMathAlphabet{\mathsfit}{bold}{\encodingdefault}{\sfdefault}{bx}{n}

%% file: figures/teaser.tex
\begin{figure}[!ht]
    \vspace{-20pt}
    \centering
    \includegraphics[width=\linewidth]{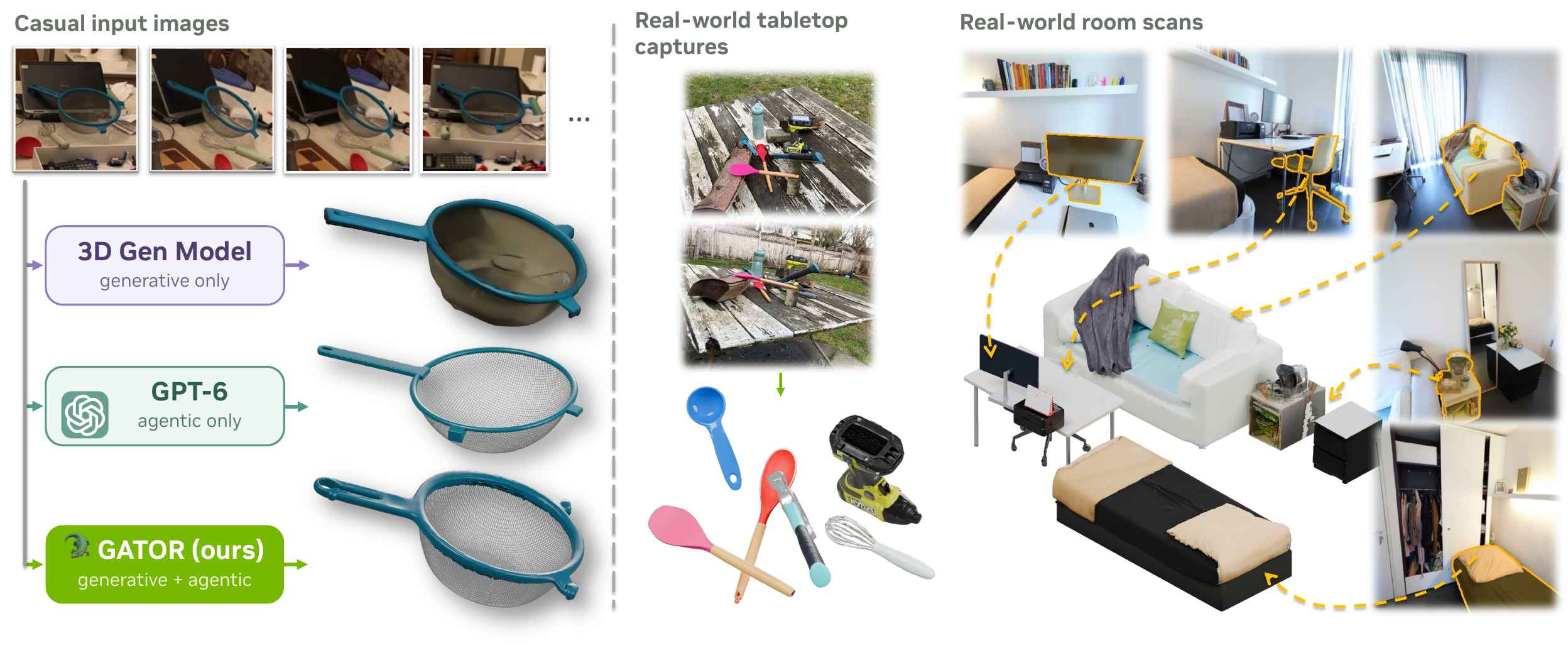}
    \vspace{-25pt}
    \caption{\textbf{Generative and agentic 3D object reconstruction from casual images.}
    \textbf{Left:} Generative model inaccurately fills the strainer's mesh bowl. GPT-6-Astra struggles with strainer ear position, and simplifies the handle and ear shapes. Our \modelname{} generates and refines the asset, preserving its shape and mesh.
    \textbf{Right:} \modelname{} reconstructs complete, textured, posed objects from real-world tabletop captures and room scans, which are composed using their predicted poses.
    \mbox{Project page: {\protect\hypersetup{hidelinks}\href{https://research.nvidia.com/labs/lpr/gator/}{\textcolor[HTML]{0057D9}{\nolinkurl{https://research.nvidia.com/labs/lpr/gator/}}}}}}
    \label{fig:teaser}
\end{figure}

%% file: sections/abstract.tex
\begin{abstract}
Reconstructing complete, scene-aligned 3D objects from casual images requires integrating sparse, uncertain observations and inferring surfaces hidden by occlusions.
We present \modelname{}, a generative and agentic framework that recovers textured object assets and their scene-relative pose from one or more images.
Our local modality mixer couples patch-aligned RGB, target-mask, and pointmap features before cross-view reasoning, preserving scene context while distinguishing the target from its surroundings.
Text-guided semantic conditioning complements these spatial cues with category names and object captions through stage-specific adapters for structure, geometry, and appearance generation.
The generated asset initializes a multimodal agent, providing instance-specific geometry and pose for targeted structural and texture refinement through an observation-guided edit-render-review loop.
Across synthetic objects, cluttered tabletops, and indoor scenes, \modelname{} achieves strong geometric and appearance fidelity while recovering scene-relative pose from sparse observations.
Time-budget comparisons and scene-level simulation further demonstrate the reconstruction efficiency and simulation readiness.
\end{abstract}

%% file: sections/introduction.tex
\section{Introduction}
\label{sec:introduction}

Turning everyday photographs into 3D assets for simulation and scene editing requires complete object geometry and appearance, together with the pose needed to place each object back into its scene.
We study \emph{casual images}: one or a few observations acquired without requiring object isolation, calibrated capture, or a prescribed camera trajectory.
The target may remain embedded in a cluttered tabletop or room, partly occluded by surrounding objects and observed from irregular viewpoints.
Lighting, apparent scale, and visible surface coverage vary across images, while additional information such as camera parameters and depth may be estimated from the observations.

Single-view object reconstruction methods infer complete 3D objects from one image.
LRM~\citep{hong2023lrm} predicts a neural radiance field, TRELLIS.2~\citep{xiang2026trellis2} generates geometry and materials, and SAM 3D~\citep{chen2026sam3d} jointly predicts object shape, texture, and layout.
However, the geometry and appearance of unobserved backsides must be inferred from learned priors, so visually plausible completions can differ from the actual object.
Exploiting additional observed views requires multiview adaptation of these single-image formulations.

Multiview generative reconstruction reduces this ambiguity by combining learned 3D priors with complementary observations.
ShapeR~\citep{siddiqui2026shaper} combines posed images, sparse SLAM points, and text captions for shape generation, ReconViaGen~\citep{chang2026reconviagen} conditions generation on multiview reconstruction features, and Pixal3D~\citep{li2026pixal3d} aggregates pixel-aligned image features in 3D.
For cluttered targets in casually captured real-world scenes, however, conditioning must associate appearance, observed geometry, and masks distinguishing the target object from surrounding objects and background, while preserving scene context.
When modalities are supplied as independent conditioning streams, the denoiser must infer these local associations.
Generation alone also offers limited explicit control over the surface regularity, part connections, and precise texturing useful for downstream applications such as simulation and gaming.

Recently, multimodal agents powered by GPT-6-Astra~\citep{openai2026gpt6astra} provide direct control over structure and materials through graphics programs and visual feedback.
However, building detailed assets from scratch requires repeated model calls, code execution, and rendering, making the process time-consuming and costly with commercial APIs.
Fine-grained geometry remains challenging: an agent may recover recognizable object structure yet reduce parts to simplified shapes (Fig.~\ref{fig:teaser}).

Addressing the aforementioned challenges requires observation-grounded completion together with explicit control over geometry and appearance.
We present \modelname{}, a generative and agentic framework for complete, pose-aware object reconstruction from casual images (Fig.~\ref{fig:teaser}).
Our key design separates patch-level cross-modal fusion from global cross-view reasoning and introduces stage-specific text guidance throughout the generative cascade.
The resulting complete, scene-aligned asset anchors agentic refinement to instance-specific geometry, appearance, and pose, enabling targeted structural and texture edits guided by the input images.

We introduce \emph{local modality mixing} into the structure-geometry-appearance cascade of TRELLIS.2~\citep{xiang2026trellis2}, fusing patch-aligned RGB, pointmap, and target-mask features before sparse structure generation.
This separates local cross-modal association from global cross-view reasoning, retaining scene context while explicitly identifying target evidence.
Once sparse structure is available, we follow Pixal3D~\citep{li2026pixal3d} in projecting image features onto its 3D locations to condition geometry and appearance generation.
Pose-aligned training preserves scene-relative orientation, while scale and placement come from the observations and are refined by differentiable rendering, so reconstructed objects can be composed in a shared scene.

Beyond the image-derived conditioning, we introduce \emph{text-guided semantic conditioning} through embeddings of category names and object captions.
These global semantic cues complement the modality mixer's patch-level spatial evidence when sparse views or occlusion leave multiple plausible object completions.
While ShapeR~\citep{siddiqui2026shaper} uses captions for shape generation, our stage-specific text adapters guide structure, geometry, and appearance, promoting category-consistent completion throughout the cascade.

The resulting complete, scene-aligned asset initializes agentic refinement, supplying instance-specific geometry, appearance, and pose rather than requiring construction from scratch.
Structural priors about man-made objects guide geometry edits, while input images guide texture refinement, including the content, placement, sharpness, and legibility of text and logos.
An edit-render-review loop compares renders of the original and modified assets under matched cameras and lighting, retaining or reverting edits based on their agreement with the observations.
The process preserves scene-relative placement and requires no retraining of the generative model.

We evaluate on Toys4K~\citep{stojanov2021toys4k} and four real-world datasets: LM-O~\citep{brachmann2014lmo}, HB~\citep{kaskman2019homebreweddb}, HANDAL~\citep{guo2023handal}, and ScanNet++~\citep{yeshwanth2023scannetpp}.
The evaluations span controlled objects, cluttered tabletops, and indoor scenes, including reconstruction from estimated masks and geometry.
\modelname{} achieves strong generalization across various challenging casual image capturing setups.
Time-budget comparisons and scene-level simulation further demonstrate the benefits of asset-initialized refinement for efficiency and stability.

Our contributions can be summarized as follows:
\begingroup
\setlength{\leftmargini}{1em}
\setlength{\topsep}{0pt}
\setlength{\partopsep}{0pt}
\setlength{\itemsep}{0pt}
\setlength{\parsep}{0pt}
\setlength{\parskip}{0pt}
\begin{itemize}
    \item \textbf{Local modality mixing for pose-aware generation.}
    We introduce a modality mixer that associates patch-aligned RGB, pointmap, and target-mask features before sparse structure generation, followed by projected image conditioning for geometry and appearance.
    This supports target-specific completion in clutter and scene-relative pose recovery from estimated inputs, allowing reconstructed objects to be composed in a shared scene.
    \item \textbf{Text-guided semantic conditioning.}
    We condition structure, geometry, and appearance generation on category names and object captions through stage-specific adapters, complementing the modality mixer's patch-level evidence with global object semantics.
    This guides completion toward category-consistent geometry and appearance when observations are sparse or occluded.
    \item \textbf{Agentic geometry and appearance refinement.}
    We initialize an observation-grounded edit-render-review loop with the generated asset, enabling an agent to apply structural and appearance priors without retraining the generative model.
    This enables targeted improvements to structural regularity and texture legibility, including text and logos, making assets better suited to downstream applications such as simulation and gaming.
    \item \textbf{Reconstruction across real-world settings.}
    Evaluations on multiple synthetic and real-world datasets cover diverse categories, sparse and incomplete view coverage, clutter and occlusion, lighting variation, and uncertainty in estimated masks, cameras, and depth.
    The results demonstrate strong geometry, appearance, and scene-relative pose recovery across these conditions.
\end{itemize}
\endgroup

%% file: sections/related_work.tex
\section{Related Work}
\label{sec:related_work}

\boldparagraphstart{Non-Generative Object Reconstruction.}
Feed-forward reconstruction models recover 3D objects from one or a few images.
LRM~\citep{hong2023lrm} and TripoSR~\citep{tochilkin2024triposr} predict radiance-field representations from single images, while SF3D~\citep{boss2024sf3d} produces textured meshes with disentangled illumination.
RaySt3R~\citep{duisterhof2025rayst3r} completes object geometry from a single masked RGB-D image through novel-view depth prediction.
For sparse multiview inputs, GRM~\citep{xu2024grm} predicts 3D Gaussians and MeshLRM~\citep{wei2024meshlrm} reconstructs meshes; LSRM~\citep{li2026lsrm} scales spatial attention to recover finer geometry and appearance.
Camera uncertainty is particularly relevant to casual capture: LEAP~\citep{jiang2023leap} aggregates views without input poses, while PF-LRM~\citep{wang2023pflrm} and FreeSplatter~\citep{xu2024freesplatter} recover both 3D representations and camera parameters from unposed images.
Geometry estimators such as VGGT~\citep{wang2025vggt} and VGGT-$\Omega$~\citep{wang2026vggtomega} provide cameras and depth for downstream reconstruction.
\modelname{} uses such estimates as geometric evidence to condition complete, pose-aware object generation.

\boldparagraphstart{Generative Object Reconstruction.}
Generative methods model plausible completions of geometry that the input images do not constrain.
Novel-view diffusion methods, including Zero-1-to-3~\citep{liu2023zero123}, SyncDreamer~\citep{liu2023syncdreamer}, Wonder3D~\citep{long2023wonder3d}, and One-2-3-45++~\citep{liu2023one2345pp}, synthesize additional image or normal observations for reconstruction.
SpaRP~\citep{xu2024sparp} jointly predicts multiview images and pose-related representations from unposed inputs, while SPAR3D~\citep{huang2025spar3d} conditions mesh reconstruction on a generated point cloud.
Native 3D generative models such as TRELLIS~\citep{xiang2024trellis} and TRELLIS.2~\citep{xiang2026trellis2} instead generate structured 3D latents, and \modelname{} adopts TRELLIS.2's structure-geometry-appearance cascade.
Aligning generation with observations is central to reconstruction: Pixal3D~\citep{li2026pixal3d} projects image features onto 3D structure, CUPID~\citep{huang2026cupid} jointly generates geometry and pixel-voxel correspondences for camera recovery, and World Tracing~\citep{zhang2026worldtracing} completes layered geometry in camera space.
ReconViaGen~\citep{chang2026reconviagen} integrates reconstruction features into a generative model, RecGen3D~\citep{huang2026recgen3d} aligns reconstruction and generation in a shared canonical space, and Mix3R~\citep{lin2026mix3r} jointly predicts sparse structure, pointmaps, and cameras.
FlowObject~\citep{rao2026flowobject} guides a pretrained flow model at inference time.
UMI3D~\citep{qu2026umi3d} and MV-SAM3D~\citep{li2026mvsam3d} adapt existing generative models to multiple images without additional training, while ASV3D~\citep{huynh2026asv3d} studies auxiliary-view adaptation.
We explicitly couple patch-aligned RGB, target-mask, and pointmap evidence before cross-view attention, with stage-specific text guidance complementing spatial conditioning throughout the generative cascade.

\boldparagraphstart{Object Reconstruction in the Wild.}
Natural captures introduce clutter, occlusion, and incomplete object coverage.
FroDO~\citep{li2020frodo}, vMAP~\citep{kong2023vmap}, and BundleSDF~\citep{wen2023bundlesdf} integrate RGB or RGB-D sequences to estimate object geometry and pose.
O$^2$-Recon~\citep{hu2023o2recon}, ROODI~\citep{chang2025roodi}, and Amodal3R~\citep{wu2025amodal3r} use learned priors to reconstruct surfaces hidden by occlusion.
Any6D~\citep{lee2025any6d} aligns generated meshes for pose and size estimation, while CARI4D~\citep{xie2026cari4d} reconstructs metric-scale human-object interactions from monocular video.
Closer to our setting, ShapeR~\citep{siddiqui2026shaper} conditions generation on casual captures and sparse SLAM points, while SAM 3D~\citep{chen2026sam3d} predicts complete textured assets from natural images.
InstaScene~\citep{yang2025instascenecomplete3dinstance}, RecGen~\citep{zadaianchuk2026recgen}, and MessyKitchens~\citep{ansari2026messykitchenscontactrichobjectlevel3d} address object reconstruction within cluttered scenes.
Our \modelname{} retains surrounding scene context while using target masks to distinguish the object from clutter, reconstructing complete geometry, appearance, and scene-relative pose from sparse, partially occluded observations.

%% file: sections/approach.tex
\section{Methodology}
\label{sec:approach}

\input{figures/model_overview}

Fig.~\ref{fig:model_overview} outlines our pipeline.
We define the reconstruction setting (Sec.~\ref{sec:problem_formulation}) and encode RGB, masks, and pointmaps on a shared patch grid (Sec.~\ref{sec:multimodal_conditioning}).
Local modality mixing then supports sparse structure generation (Sec.~\ref{sec:sparse_structure_generation}), followed by geometry and appearance generation from projected image features (Sec.~\ref{sec:geometry_appearance}), with text guiding the whole cascade.
The resulting scene-aligned asset initializes an agent that uses the input images and rendered feedback for refinement (Sec.~\ref{sec:agent_refinement}).

\subsection{Problem Formulation}
\label{sec:problem_formulation}

Given $N$ casual images of a target object, we generate its complete geometry and appearance in a shared reconstruction frame.
Each view provides RGB $I_v$, a target mask $M_v$, intrinsics $K_v$, camera pose $T_v$, and depth $D_v$, all of which can be estimated from the images rather than given as ground truth.
With text $c$, such as a category name or a VLM caption, the conditioning set is $\mathcal{C}=(\{I_v,M_v,K_v,T_v,D_v\}_{v=1}^{N},c)$.
Unlike RecGen~\citep{zadaianchuk2026recgen}, which accepts at most two views, \modelname{} takes any number of views and improves with more (Sec.~\ref{sec:ablation_studies}).

The center and extent of the masked depth, unprojected to 3D, define an observation anchor $\mathbf{t}_g$ and an isotropic scale $s_g$.
We normalize observations by this anchor without rotating the frame and map generated points back by $\mathbf{x}=s_g\mathbf{q}+\mathbf{t}_g$, so outputs keep the scene-relative orientation of the observations (Sec.~\ref{app:normalization}).
The anchor only initializes translation and scale, which differentiable rendering refines after generation (Sec.~\ref{sec:geometry_appearance}).

\subsection{Multimodal Observation Encoding}
\label{sec:multimodal_conditioning}

\modelname{} encodes RGB, the target mask, and a pointmap unprojected from depth on a shared patch grid of consistently cropped views, using frozen DINOv3 features~\citep{simeoni2025dinov3} for RGB and masks and a trainable pointmap encoder.
RGB and pointmaps keep surrounding context, while mask tokens identify the target.
ShapeR~\citep{siddiqui2026shaper} uses similar inputs but relies on SLAM points pre-segmented for the target, which images alone do not provide, and combines separately encoded tokens only inside its denoiser.
RecGen~\citep{zadaianchuk2026recgen} fuses earlier but simply adds mask and pointmap embeddings to its image tokens.
Our modality mixer instead applies attention among the aligned RGB, mask, and pointmap tokens at each patch within a view, weighting the three cues according to the patch content while keeping a token per modality for cross-view aggregation.
This encodes known spatial correspondence into the fusion architecture instead of leaving the structure denoiser to infer it (Sec.~\ref{sec:sparse_structure_generation}).
Text also conditions every stage, modulating structure generation globally and entering geometry and appearance as tokens (Sec.~\ref{sec:geometry_appearance}), whereas ShapeR adds text tokens only to its shape denoiser.

\subsection{Aligning and Mixing Modalities for Structure Generation}
\label{sec:sparse_structure_generation}

For patch $p$ in view $v$, let $\mathbf{h}^{m}_{vp}$ be the token of modality $m\in\{I,M,P\}$, and let $\mathbf{a}_{vp}=A(\gamma(\mathbf{u}_p),\mathbf{r}_{vp})$ embed its Fourier-encoded patch coordinates and patch-averaged Pl\"ucker ray.
The modality mixer applies residual attention across the three tokens of each patch:
\begin{equation}
    \{\widetilde{\mathbf{h}}^{m}_{vp}\}_{m\in\{I,M,P\}}
    =\operatorname{ModalityMixer}\!\left(
      \{\mathbf{h}^{m}_{vp}+\mathbf{a}_{vp}\}_{m\in\{I,M,P\}}
    \right).
    \label{eq:modality_mixer}
\end{equation}
Each output token keeps its modality identity but can combine appearance, geometry, and mask cues that separate the target from its surroundings.
Seeing all three cues of a patch together also lets the model cross-check estimated depth against appearance and the mask, tolerating upstream errors such as noisy boundary depth or imprecise mask edges.

The dense structure latents then query the bank $\mathcal{B}$ of mixed tokens from all views through cross-attention.
Because the sparse structure is not yet known, this global attention, rather than local projection, gives each latent access to every view and its context, allowing completion under limited coverage and occlusion.
Pixal3D~\citep{li2026pixal3d}, in contrast, implicitly assumes that the target is fully visible from its anchor view, making it more brittle to occlusion or partial observations.
A pooled SigLIP2 embedding~\citep{tschannen2025siglip2} of the category text, passed through a learned adapter, modulates each block via the timestep embedding and adaptive layer normalization as a semantic prior.
The denoiser learns rectified flow over complete, pose-aware occupancy latents, which the frozen decoder converts into sparse structure $\widehat{\mathcal{S}}$ (Secs.~\ref{app:observation_encoding} and~\ref{app:flow_matching}).

\subsection{Structure and Text Conditioned Geometry and Appearance}
\label{sec:geometry_appearance}

Given the sparse structure, we lift image features to its active locations.
For each location $\mathbf{q}_j$, we average frozen DINOv3 features and their NAF-upsampled versions~\citep{chambon2025naf}, sampled from the target-masked images:
\begin{equation}
    \mathbf{f}_j=\frac{1}{|\mathcal{V}_j|}
    \sum_{v\in\mathcal{V}_j}
    [F_v^{D}(\pi_v(s_g\mathbf{q}_j+\mathbf{t}_g));
     F_v^{N}(\pi_v(s_g\mathbf{q}_j+\mathbf{t}_g))],
    \label{eq:sparse_feature_projection}
\end{equation}
where $\pi_v$ projects into cropped view $v$ and $\mathcal{V}_j$ contains the views in which $\mathbf{q}_j$ lies in front of the camera and projects inside the crop ($\mathbf{f}_j=\mathbf{0}$ if empty).
In each denoising block, cross-attention to global image and text tokens guides unobserved regions, and a learned projection of $\mathbf{f}_j$ is then added for local detail.
The geometry flow generates shape latents, and the appearance flow generates physically based rendering (PBR) material latents at the same sparse coordinates, conditioned on those shape latents. Both flows are trained on ground-truth sparse structure and use the predicted sparse structure at inference.
A final placement step refines translation and isotropic scale by differentiable rendering of silhouettes and depth and by point alignment, keeping the predicted orientation and shape (Sec.~\ref{app:feature_projection}).
Compared with Pixal3D, which also projects image features onto sparse structure, our design generates that structure from mixed RGB, mask, and pointmap tokens of all views, adds text guidance, and predicts scene placement.

\boldparagraphstart{Text conditioning.}
Prior image-conditioned generative models~\citep{xiang2026trellis2,li2026pixal3d,chen2026sam3d} often ignore text information which helps to guide plausible completions when views are sparse or occluded. Therefore, we condition every stage on text: besides the SigLIP2 modulation of structure generation, a frozen T5Gemma2 encoder~\citep{zhang2025t5gemma2} and a learned adapter produce text tokens for the geometry and appearance flows.
While structure generation takes a short category prompt, these flows are trained on captions ranging from brief high-level category names to detailed long descriptions to provide more nuanced guidance.

\boldparagraphstart{Masks across stages.}
Masks play a different role in each stage, which matters in cluttered casual captures.
The structure flow sees the full crop, including occluders and supporting surfaces, with the mask as separate tokens, so it can account for occlusion when completing the object.
The geometry and appearance flows instead project features from target-masked images, so occluders and background do not leak into surface detail or texture.
By contrast, TRELLIS.2, Pixal3D, and ReconViaGen condition every stage on background-removed images~\citep{xiang2026trellis2,li2026pixal3d,chang2026reconviagen}, discarding the scene context that helps complete occluded objects.

\subsection{Agent as Reconstruction Refiner}
\label{sec:agent_refinement}

The generated asset can still have missing parts, fused openings, or texture contamination.
We provide GPT-6-Astra~\citep{openai2026gpt6astra}, a multimodal agent, with our generated mesh, input observations, and Blender access. It is prompted to make targeted repairs that preserve reliable regions and the asset's coordinate frame and scale rather than remodel the object.
The agent decides what to fix, inspecting the images and asset renders and writing scripts to complete missing structure, repair topology, or correct UVs and materials, guided by the mesh's proportions, orientation, and attachment sites.
After each edit, it compares renders of the original and edited assets under matched cameras and lighting and reverts edits that degrade the reconstruction.
It stops when supported repairs are complete or evidence is insufficient for further edits, within a fixed time budget (Sec.~\ref{app:agent_refinement}).

%% file: figures/model_overview.tex
\begin{figure}[!t]
    \centering
    \includegraphics[width=\linewidth]{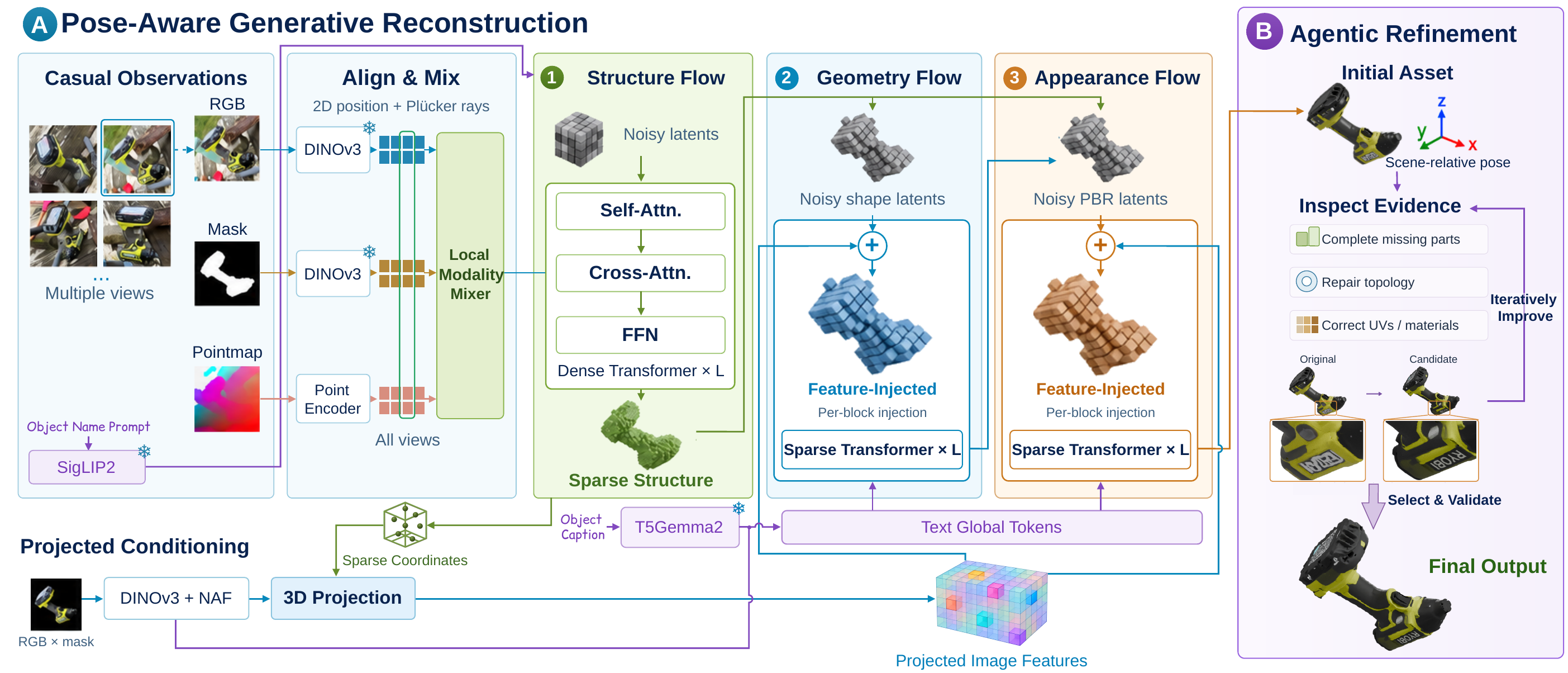}
    \vspace{-20pt}
    \caption{\textbf{\modelname{} architecture.}
    \textbf{(A) Pose-aware generative reconstruction.}
    Following TRELLIS.2~\citep{xiang2026trellis2}, \modelname{} uses conditional flows for sparse structure, geometry, and appearance.
    RGB images and depth-derived pointmaps retain scene context, while masks identify the target. Tokens from all three modalities share a patch grid with common 2D position and Pl\"ucker-ray encodings. A local mixer exchanges information at each patch within a view (Secs.~\ref{sec:multimodal_conditioning} and~\ref{sec:sparse_structure_generation}).
    Cross-view attention and SigLIP2 object-name conditioning guide structure generation.
    At its active coordinates, projected DINOv3 and NAF-upsampled features from target-masked images are averaged across valid views and injected into every geometry and appearance denoising block.
    Global image tokens and T5Gemma2-encoded captions supply cross-attention. Shape latents condition PBR material generation (Sec.~\ref{sec:geometry_appearance}).
    \textcolor[HTML]{2785AC}{\ding{100}} denotes frozen encoders.
    \textbf{(B) Agentic refinement.}
    Initialized with the complete, scene-aligned asset, an agent uses input images and structural priors to complete parts, repair topology, and refine texture coordinates and materials. Its edit-render-review loop compares original and candidate renders under matched cameras and lighting to retain or revert local edits. Within the time budget, it selects the best inspected asset while preserving reliable regions and scene-relative placement (Sec.~\ref{sec:agent_refinement}).}
    \label{fig:model_overview}
    \vspace{-15pt}
\end{figure}

%% file: sections/experiments.tex
\section{Experiments}
\label{sec:experiments}

\subsection{Experimental Setup}
\label{sec:experimental_datasets}
\label{sec:evaluation_metrics}
\label{sec:synthetic_scene_training}

\boldparagraphstart{Training data.}
We train only on synthetic scenes: procedurally cluttered scenes composed from approximately 980,000 object assets, plus structured indoor scenes from existing datasets~\citep{fu2021front,zhong2025internscenes,pfaff2026scenesmithagenticgenerationsimulationready,xia2026sagescalableagentic3d}.
Each example pairs partial observations of a target with its complete, posed geometry and appearance, with observations augmented or replaced by image-estimated cameras and depth to mimic casual capture (Sec.~\ref{app:datasets}).

\boldparagraphstart{Evaluation data.}
To study complete object reconstruction and  placement pose in the original scene from sparse images, we evaluate on 525 Toys4K~\citep{stojanov2021toys4k} objects from 105 categories and
real world captures including  LM-O~\citep{brachmann2014lmo}, HB~\citep{kaskman2019homebreweddb}, HANDAL~\citep{guo2023handal}, and ScanNet++~\citep{yeshwanth2023scannetpp}.  These diverse datasets extensively cover various settings including controlled object renderings, cluttered tabletops, and indoor scans.
LM-O and HB use sensor depth and recovered camera poses, while HANDAL and ScanNet++ use estimated cameras and depth. Masks come from the datasets except on ScanNet++, where they are inferred by SAM3~\citep{carion2026sam3}.
All methods share the same input setup (Secs.~\ref{app:toys4k_mv4}, \ref{app:tabletop}, and~\ref{app:scannetpp}).
Our refinement and the from-scratch GPT-6-Astra baseline both use GPT-6-Astra~\citep{openai2026gpt6astra} with extra-high reasoning effort and a ten-minute budget per object unless otherwise mentioned. 
Implementation details are in Sec.~\ref{app:implementation}.

\boldparagraphstart{Metrics.}
Joint pose and shape are measured by bidirectional posed-surface distance (ADD-SB) and its recall \addsbrecall{} at 10\% of the reference diameter~\citep{chen2026sam3d,zadaianchuk2026recgen}.
After similarity alignment to ground truth, we report chamfer distance (CD), normal consistency (NC), F1 at 1\% (Toys4K) or 2\% (real data) of the diameter, and 2D silhouette IoU~\citep{li2026pixal3d}.
Appearance uses PSNR, SSIM, and LPIPS~\citep{zhang2018lpips} on renders under the same alignment and shared lighting.
All distances are normalized by the reference diameter (Sec.~\ref{app:metrics}).

\input{tables/toys4k_results}

\subsection{Toys4K Results}
\label{sec:toys4k_results}

Table~\ref{tab:toys4k_results} compares against TRELLIS.2~\citep{xiang2026trellis2}, ReconViaGen~\citep{chang2026reconviagen}, Pixal3D~\citep{li2026pixal3d} (single- and multi-view checkpoints), and GPT-6-Astra from scratch, all using four views. \modelname{}, Pixal3D, and GPT-6-Astra share Pi3X~\citep{wang2025pi3} camera and depth estimates.
\modelname{} leads all seven metrics, reducing CD by 19.7\% and raising F1 from 0.701 to 0.757 relative to the strongest baseline, Pixal3D-MV.
Qualitative comparisons appear in Sec.~\ref{app:toys4k_mv4}.

\subsection{Real-World Results}
\label{sec:real_world_results}
\label{sec:real_multiview_results}
\label{sec:handal_scannetpp_results}

\input{tables/real_world_mv4_results}

\boldparagraphstart{Tabletop objects.}
Table~\ref{tab:real_world_mv4_results} covers all target objects from LM-O, HB, and HANDAL datasets. For each object, four-views are provided as input. We compare against ReconViaGen~\citep{chang2026reconviagen}, Pixal3D~\citep{li2026pixal3d}, MV-SAM3D~\citep{li2026mvsam3d}, and ShapeR~\citep{siddiqui2026shaper}, RecGen~\citep{zadaianchuk2026recgen}, SimFoundry~\citep{ranawaka2026simfoundry}, and GPT-6-Astra from scratch, where RecGen and SimFoundry selects the best two and one views respectively from the given input views, due to their input view limits.
Excluding the fully proprietary GPT-6-Astra baseline, \modelname{} leads all nine metrics, with its largest margins in geometry: it roughly halves Pixal3D's CD, while drastically raising F1 and PSNR (Fig.~\ref{fig:real_tabletop}).

\input{figures/real_tabletop}

\input{tables/handal_scannetpp_results}

\boldparagraphstart{Objects in indoor scans.}
Table~\ref{tab:handal_scannetpp_results} extends the ScanNet++ evaluation protocol~\citep{ni2025decompositionalneuralscenereconstruction,siddiqui2026shaper,wu2026jrmjointreconstructionmodel} from 69 objects in 6 scenes to 229 targets from 49 scenes, each with one to ten views.
Despite estimated cameras, depth, and masks, \modelname{} leads every metric, reducing CD by 23\% relative to the runner-up, RecGen, and raising F1 from 0.540 to 0.603 and IoU from 0.682 to 0.733.
It also surpasses GPT-6-Astra's from-scratch reconstructions on every metric, so its advantage carries over from tabletop captures to objects in cluttered indoor scenes.
Figs.~\ref{fig:real_scannetpp} and~\ref{fig:real_scannetpp_additional} show qualitative comparisons.

\input{figures/real_scannetpp}

\boldparagraphstart{Pose-aware reconstruction.}
Measured before any object-to-reference alignment to account for pose estimation from each method, \modelname{} reaches \addsbrecall{} of 96.36\% on HANDAL and 95.63\% on ScanNet++, and the lowest mean ADD-SB on both the tabletop and room-level ScanNet++ benchmarks.

\subsection{Framework Analysis}
\label{sec:ablation_studies}

\input{tables/ablation_results}

\boldparagraphstart{Ablation study.}
Table~\ref{tab:ablation_results} adds one component at a time on 55 HANDAL objects with four input views. All variants are trained on the same synthetic data with the same pretrained image encoders and share observations and sampling settings (Sec.~\ref{app:ablation}).
The modality mixer mainly improves placement including both geometry and pose estimation by  lowering ADD-SB. Since camera and depth information are inferred by external models, this suggests the mixer enhances robustness to upstream estimation errors.
Text conditioning then improves metrics consistently, indicating that category semantics help complete shapes under challenging casual image captures. Finally, agent refinement generally enhances appearance (Table~\ref{tab:ablation_results}) and on every tabletop dataset (Table~\ref{tab:real_world_mv4_by_dataset}).
Its edits are local, such as the reopened holes and slots in Fig.~\ref{fig:ablation_variants}, so average geometry changes little on Toys4K, LM-O, and HB and only slightly worsens CD, F1, and IoU on HANDAL. However, it benefits downstream applications such as simulation (Sec.~\ref{app:sim_readiness}), which cannot be fully reflected in the quantitative metrics.

\input{figures/efficiency}

\boldparagraphstart{Efficiency analysis.}
Fig.~\ref{fig:efficiency_analysis}(a) plots quality against mean pipeline work, with \modelname{} adding an estimated 32\,s for our generative initialization.
Without refinement, \modelname{} takes about 0.5\,min and already outperforms GPT-6-Astra run with a 20-minute budget.
Refinement then mainly improves appearance and local topology.
GPT-6-Astra returns valid assets for 0/10 objects at a one-minute budget, 5/10 at two minutes, and all ten from five minutes onward.
The example in panel (b) illustrates the difficulty of constructing a complete asset from scratch under short budgets.

\boldparagraphstart{Effect of input views.}
Our single model supports varying numbers of views and improves steadily as views are added: on LM-O and HB (Table~\ref{tab:bop_multiview_results}), going from two to eight views lowers ADD-SB  and CD. Even with two views its CD is about half of Pixal3D's with eight.

%% file: tables/toys4k_results.tex
\begin{table}[!tb]
    \centering
    \small
    \setlength{\tabcolsep}{2pt}
    \setlength{\aboverulesep}{0pt}
    \setlength{\belowrulesep}{0pt}
    \renewcommand{\arraystretch}{1.12}
    \caption{\textbf{Four-view reconstruction on Toys4K.}
    $^\dagger$Run with MultiDiffusion~\citep{bartal2023multidiffusion}.
    Dark/light green mark best/second-best values (excluding the fully proprietary GPT-6-Astra baseline).}
    \label{tab:toys4k_results}
    \vspace{-8pt}
    \scalebox{0.9}{%
    \begin{tabularx}{\linewidth}{@{}l *{7}{>{\centering\arraybackslash}X}}
        \toprule
        \multirow{2}{*}{Methods}
        & \multicolumn{4}{c}{Geometric}
        & \multicolumn{3}{c}{Appearance} \\
        \cmidrule(lr){2-5}\cmidrule(lr){6-8}
        & CD$\downarrow$ & NC$\uparrow$ & F1$\uparrow$ & IoU$_{2\mathrm{D}}\uparrow$
        & PSNR$\uparrow$ & SSIM$\uparrow$ & LPIPS$\downarrow$ \\
        \midrule
        \rowcolor{black!10}
        GPT-6-Astra & 0.013 & 0.754 & 0.614 & 0.774 & 17.11 & 0.865 & 0.214 \\
        TRELLIS.2$^\dagger$ & 0.014 & 0.751 & 0.628 & 0.747 & 15.73 & 0.855 & 0.231 \\
        ReconViaGen & 0.014 & 0.767 & 0.579 & 0.764 & 14.39 & 0.825 & 0.228 \\
        Pixal3D-SV & 0.027 & 0.681 & 0.433 & 0.629 & 14.22 & 0.844 & 0.287 \\
        Pixal3D-MV & \secondresult{0.011} & \secondresult{0.783} & \secondresult{0.701} & \secondresult{0.798} & \secondresult{17.18} & \secondresult{0.871} & \secondresult{0.203} \\
        \modelname{} (Ours) & \bestresult{0.009} & \bestresult{0.801} & \bestresult{0.757} & \bestresult{0.819} & \bestresult{17.60} & \bestresult{0.873} & \bestresult{0.191} \\
        \bottomrule
    \end{tabularx}%
    }
\end{table}

%% file: tables/real_world_mv4_results.tex
\begin{table}[!tb]
    \centering
    \footnotesize
    \setlength{\tabcolsep}{1pt}
    \setlength{\aboverulesep}{0pt}
    \setlength{\belowrulesep}{0pt}
    \renewcommand{\arraystretch}{1.12}
    \caption{\textbf{Real-world tabletop reconstruction on LM-O, HB, and HANDAL.}
    ReconViaGen and Pixal3D lack scene placement, so joint ADD-SB is unavailable.
    ShapeR can only output untextured object meshes, so appearance scores are unavailable.
    Dark/light green mark best/second-best scores, excluding the fully proprietary GPT-6-Astra baseline.
    \modelname{} leads all metrics among the remaining methods, roughly halving Pixal3D's chamfer distance (per-dataset results: Table~\ref{tab:real_world_mv4_by_dataset}).
}
    \label{tab:real_world_mv4_results}
    \vspace{-7pt}
    \scalebox{0.9}{%
    \begin{tabularx}{\linewidth}{l *{9}{>{\centering\arraybackslash}X}}
        \toprule
        \multirow{2}{*}{Methods}
        & \multicolumn{2}{c}{Joint (ADD-SB)}
        & \multicolumn{4}{c}{Geometry}
        & \multicolumn{3}{c}{Appearance} \\
        \cmidrule(lr){2-3}\cmidrule(lr){4-7}\cmidrule(lr){8-10}
        & Mean$\downarrow$ & @0.1$\uparrow$
        & CD$\downarrow$ & NC$\uparrow$ & F1$\uparrow$ & IoU$_{2\mathrm{D}}\!\uparrow$
        & PSNR$\uparrow$ & SSIM$\uparrow$ & LPIPS$\downarrow$ \\
        \midrule
        \rowcolor{black!10}
        GPT-6-Astra & 0.027 & 100.00 & 0.016 & 0.799 & 0.773 & 0.791 & 17.16 & 0.886 & 0.160 \\
        SimFoundry & 0.183 & 63.36 & 0.040 & 0.716 & 0.533 & 0.626 & 14.90 & 0.875 & 0.200 \\
        RecGen & 0.057 & 85.50 & 0.021 & \secondresult{0.791} & 0.666 & 0.757 & 16.00 & \secondresult{0.885} & 0.178 \\
        ReconViaGen & --- & --- & 0.027 & 0.750 & 0.608 & 0.730 & 14.48 & 0.858 & 0.181 \\
        ShapeR & 0.057 & 87.02 & 0.028 & 0.709 & 0.606 & 0.700 & --- & --- & --- \\
        MV-SAM3D & \secondresult{0.044} & \secondresult{91.60} & 0.019 & 0.778 & 0.687 & 0.757 & 15.51 & 0.883 & 0.168 \\
        Pixal3D & --- & --- & \secondresult{0.018} & 0.784 & \secondresult{0.780} & \secondresult{0.813} & \secondresult{16.94} & \bestresult{0.890} & \secondresult{0.163} \\
        \modelname{} (Ours) & \bestresult{0.022} & \bestresult{98.47} & \bestresult{0.009} & \bestresult{0.859} & \bestresult{0.904} & \bestresult{0.841} & \bestresult{17.97} & \bestresult{0.890} & \bestresult{0.140} \\
        \bottomrule
    \vspace{-20pt}

    \end{tabularx}%
    }
\end{table}

%% file: figures/real_tabletop.tex
\begin{figure}[!tb]
    \centering
    \includegraphics[width=\linewidth]{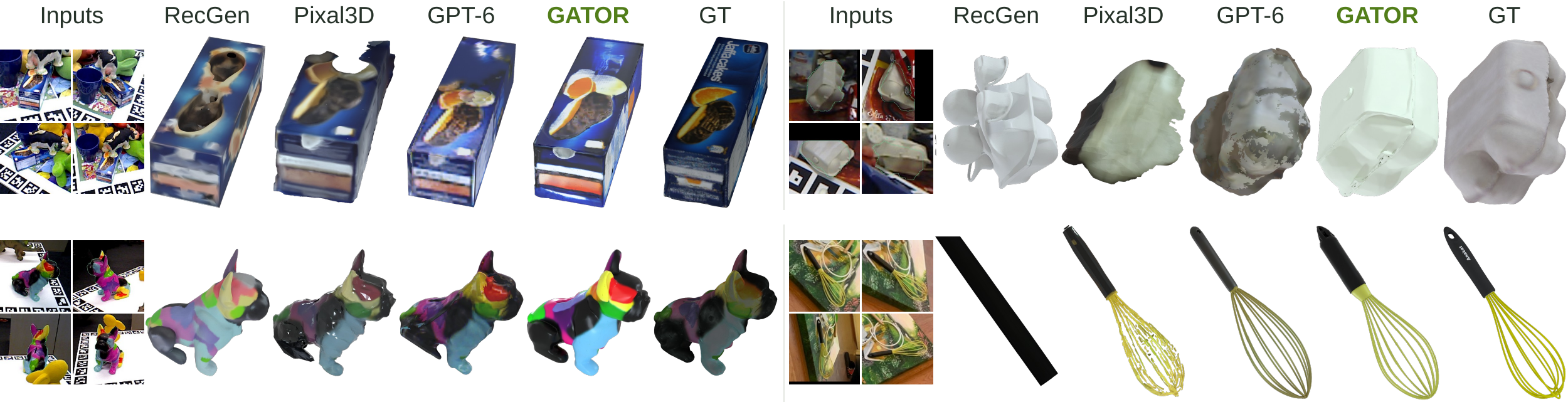}
    \vspace{-20pt}
    \caption{\textbf{Real-world tabletop reconstruction} on HB, LM-O, and HANDAL from four input views (two for RecGen); GPT-6 reconstructs from scratch.
    \modelname{} recovers complete shapes and detailed textures, such as the printed box and the whisk's wires, where baselines truncate, blur, or fail.}
    \label{fig:real_tabletop}
    \vspace{-10pt}
\end{figure}

%% file: tables/handal_scannetpp_results.tex
\begin{table}[!tb]
    \centering
    \small
    \setlength{\tabcolsep}{2pt}
    \setlength{\aboverulesep}{0pt}
    \setlength{\belowrulesep}{0pt}
    \renewcommand{\arraystretch}{1.12}
    \caption{\textbf{Real-world indoor-scene reconstruction on ScanNet++.}
    ReconViaGen and Pixal3D predict no pose placement and thus have no joint (ADD-SB) result.
    Dark/light green mark best/second-best values (excluding the fully proprietary GPT-6-Astra baseline).}
    \label{tab:handal_scannetpp_results}
    \vspace{-8pt}
    \scalebox{0.9}{%
    \begin{tabularx}{\linewidth}{l *{6}{>{\centering\arraybackslash}X}}
        \toprule
        \multirow{2}{*}{Methods}
        & \multicolumn{2}{c}{Joint (ADD-SB)}
        & \multicolumn{4}{c}{Geometry} \\
        \cmidrule(lr){2-3}\cmidrule(lr){4-7}
        & Mean$\downarrow$ & @0.1$\uparrow$
        & CD$\downarrow$ & NC$\uparrow$ & F1$\uparrow$ & IoU$_{2\mathrm{D}}\uparrow$ \\
        \midrule
        \rowcolor{black!10}
        GPT-6-Astra & 0.049 & 90.83 & 0.033 & 0.716 & 0.581 & 0.716 \\
        SimFoundry & 0.125 & 72.49 & 0.104 & 0.664 & 0.426 & 0.614 \\
        RecGen & 0.067 & \secondresult{90.83} & \secondresult{0.041} & \secondresult{0.703} & \secondresult{0.540} & \secondresult{0.682} \\
        ReconViaGen & --- & --- & 0.161 & 0.676 & 0.423 & 0.498 \\
        ShapeR & 0.071 & 81.66 & 0.060 & 0.684 & 0.458 & 0.610 \\
        MV-SAM3D & \secondresult{0.057} & \secondresult{90.83} & 0.042 & 0.699 & 0.493 & 0.648 \\
        Pixal3D & --- & --- & 0.052 & 0.679 & 0.475 & 0.623 \\
        \modelname{} (Ours) & \bestresult{0.041} & \bestresult{95.63} & \bestresult{0.031} & \bestresult{0.726} & \bestresult{0.603} & \bestresult{0.733} \\
        \bottomrule
    \end{tabularx}%
    }
    \vspace{-8pt}
\end{table}

%% file: figures/real_scannetpp.tex
\begin{figure}[!tb]
    \centering
    \includegraphics[width=\linewidth]{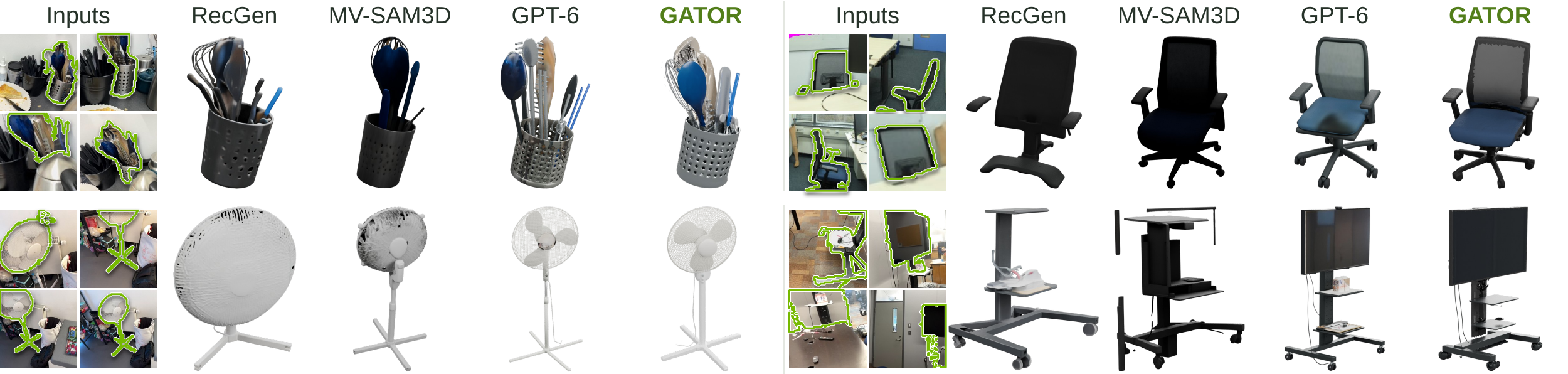}
    \vspace{-15pt}
    \caption{\textbf{Reconstruction from real-world ScanNet++ images.}
    Green outlines mark targets in a subset of the input views, and meshes are shown in comparable orientations; GPT-6 reconstructs from scratch.
    Our \modelname{} recovers fine structures, including individual utensils, fan blades, and the cart-mounted display.}
    \label{fig:real_scannetpp}
    \vspace{-15pt}
\end{figure}

%% file: tables/ablation_results.tex
\begin{table}[!tb]
    \centering
    \footnotesize
    \setlength{\tabcolsep}{1pt}
    \setlength{\aboverulesep}{0pt}
    \setlength{\belowrulesep}{0pt}
    \renewcommand{\arraystretch}{1.12}
    \caption{\textbf{Ablation on 55 HANDAL objects with four input views.}
    Each row adds one component to the row above.
    The modality mixer lowers ADD-SB, text conditioning improves all nine metrics, and agentic refinement further improves CD, NC, F1, PSNR, and LPIPS.
    }
    \label{tab:ablation_results}
    \vspace{-8pt}
    \scalebox{0.95}{%
    \begin{tabularx}{\linewidth}{l *{9}{>{\centering\arraybackslash}X}}
        \toprule
        \multirow{2}{*}{Configuration}
        & \multicolumn{2}{c}{Joint (ADD-SB)}
        & \multicolumn{4}{c}{Geometry}
        & \multicolumn{3}{c}{Appearance} \\
        \cmidrule(lr){2-3}\cmidrule(lr){4-7}\cmidrule(lr){8-10}
        & Mean$\downarrow$ & @0.1$\uparrow$ & CD$\downarrow$ & NC$\uparrow$
        & F1$\uparrow$ & IoU$_{2\mathrm{D}}\!\uparrow$
        & PSNR$\uparrow$ & SSIM$\uparrow$ & LPIPS$\downarrow$ \\
        \midrule
        Base model & 0.057 & 94.55 & 0.010 & 0.813 & 0.877 & 0.688 & 18.20 & 0.922 & 0.116 \\
        + Modality mixer & 0.037 & 94.55 & 0.011 & 0.811 & 0.872 & 0.700 & 18.16 & 0.922 & 0.114 \\
        + Text conditioning & 0.032 & \textbf{98.18} & 0.008 & 0.828 & 0.930 & \textbf{0.739} & 18.93 & \textbf{0.925} & 0.102 \\
        + Agentic refinement & \textbf{0.032} & \textbf{98.18} & \textbf{0.007} & \textbf{0.829} & \textbf{0.936} & 0.717 & \textbf{19.60} & 0.918 & \textbf{0.097} \\
        \bottomrule
    \end{tabularx}
    }
    \par\vspace{6pt}
    \input{figures/ablation_variants}
\end{table}

%% file: figures/ablation_variants.tex
\begin{minipage}{\linewidth}
    \normalsize
    \renewcommand{\arraystretch}{1}
    \captionsetup{type=figure}
    \centering
    \input{figures/ablation_variants/panel}
    \vspace{-8pt}
    \caption{\textbf{Qualitative ablation on HANDAL.}
    Columns follow the configurations in Table~\ref{tab:ablation_results}; two of the four input views are shown. Predictions are rigidly aligned to GT for shape comparison.
    Agentic refinement recovers regular holes and slots that the generative model variants leave irregular or missing.}
    \label{fig:ablation_variants}
    \vspace{-10pt}
\end{minipage}

%% file: figures/ablation_variants/panel.tex
\begingroup
\input{figures/ablation_variants/gt_oriented_crops}
\input{figures/ablation_variants/input_crops}\setlength{\tabcolsep}{0.5pt}
\newlength{\ablationgroupwidth}
\newlength{\ablationcellwidth}
\setlength{\ablationgroupwidth}{\dimexpr(\linewidth-10pt)/2\relax}
\setlength{\ablationcellwidth}{\dimexpr(\ablationgroupwidth-5pt)/6\relax}
\newcommand{\ablationrender}[3]{%
  \edef\ablationoptions{width=\the\ablationcellwidth,height=1.59\ablationcellwidth,keepaspectratio,trim=\csname ablation#1Trim\endcsname,clip}%
  \expandafter\includegraphics\expandafter[\ablationoptions]{figures/ablation_variants/assets/gt_oriented/#2/#3.png}%
}
\newcommand{\ablationphoto}[3]{%
  \edef\ablationoptions{width=0.7\ablationcellwidth,trim=\csname ablation#1Input#3Trim\endcsname,clip}%
  \expandafter\includegraphics\expandafter[\ablationoptions]{figures/ablation_variants/assets/#2/input#3.png}%
}
\newcommand{\ablationinputs}[2]{%
  \shortstack{\ablationphoto{#1}{#2}{0}\\[0.5pt]\ablationphoto{#1}{#2}{1}}%
}
\newcommand{\ablationheader}[1]{\fontsize{6.5}{7.2}\selectfont\mbox{#1}}
\newcommand{\ablationexample}[2]{%
  \begin{minipage}[t]{\ablationgroupwidth}\centering
    \begin{tabular}[t]{@{}*{6}{>{\centering\arraybackslash}m{\ablationcellwidth}}@{}}
      \ablationheader{Inputs}
      & \ablationheader{Base}
      & \ablationheader{+ Mixer}
      & \ablationheader{+ Text}
      & \ablationheader{+ Agent}
      & \ablationheader{GT}\\[2pt]
      \ablationinputs{#1}{#2}
      & \ablationrender{#1}{#2}{a01}
      & \ablationrender{#1}{#2}{a11}
      & \ablationrender{#1}{#2}{a2}
      & \ablationrender{#1}{#2}{a2ao}
      & \ablationrender{#1}{#2}{gt}
    \end{tabular}%
  \end{minipage}%
}
\noindent\ablationexample{Left}{handal_scene_000001_obj34}%
\hfill
\ablationexample{Right}{handal_scene_000009_obj10}%
\endgroup

%% file: figures/ablation_variants/gt_oriented_crops.tex
\newcommand{\ablationLeftTrim}{212bp 22bp 212bp 21bp}
\newcommand{\ablationRightTrim}{355bp 21bp 355bp 21bp}

%% file: figures/ablation_variants/input_crops.tex
\expandafter\def\csname ablationLeftInput0Trim\endcsname{248bp 56bp 55bp 121bp}
\expandafter\def\csname ablationLeftInput1Trim\endcsname{198bp 62bp 156bp 166bp}
\expandafter\def\csname ablationRightInput0Trim\endcsname{291bp 157bp 164bp 172bp}
\expandafter\def\csname ablationRightInput1Trim\endcsname{178bp 171bp 137bp 18bp}

%% file: figures/efficiency.tex
\begin{figure}[!tb]
    \centering
    \includegraphics[width=\linewidth]{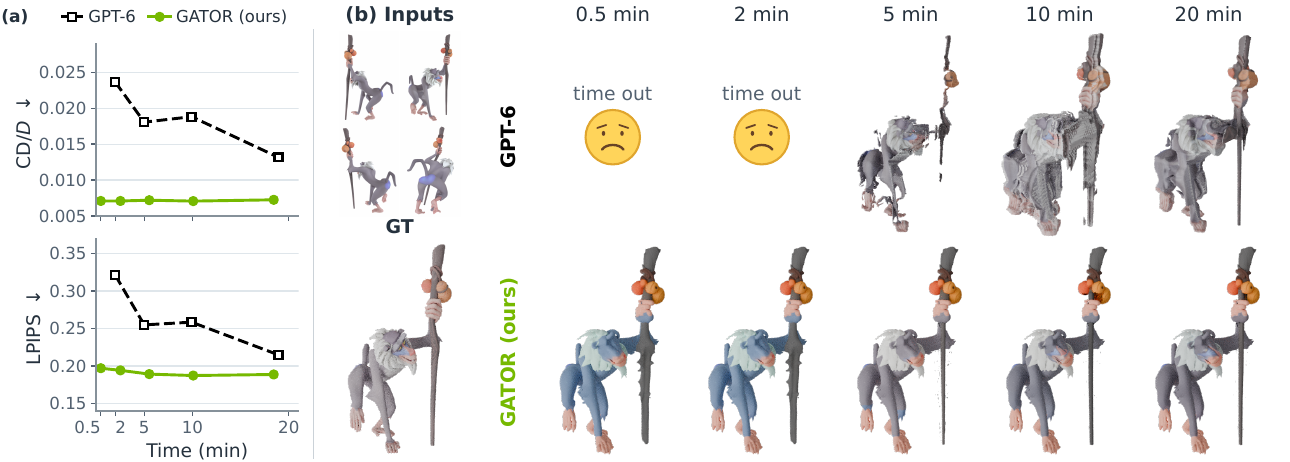}
    \vspace{-15pt}
    \caption{\textbf{Reconstruction quality versus time.}
    \textbf{(a)} Mean CD/$D$ and LPIPS ($\downarrow$) over valid outputs on sampled Toys4K objects.
    Time is the mean recorded full pipeline time, with an estimated 32\,s of generation per object for \modelname{}, and excludes mesh export.
    The first \modelname{} point is generation without refinement (about 0.5\,min), which already outperforms \mbox{GPT-6} run with a 20-minute budget.
    \textbf{(b)} One example rendered from a shared camera.
    The first column shows our result without refinement, and the remaining headings denote agent budgets.
    GPT-6 times out within 2\,min (sad faces) and remains coarse even at 20\,min.}
    \label{fig:efficiency_analysis}
    \label{fig:efficiency_qualitative}
    \vspace{-15pt}
\end{figure}

%% file: sections/conclusion.tex
\section{Conclusion}
\label{sec:conclusion}

We presented \modelname{}, a framework for complete, scene-aligned 3D object reconstruction from casual images.
Our \emph{local modality mixing} associates patch-aligned RGB, pointmap, and target-mask evidence before cross-view reasoning, preserving context while identifying the target.
Complementary \emph{text-guided semantic conditioning} supplies category priors across structure, geometry, and appearance generation to guide completion from sparse or occluded observations.
The generated asset anchors agentic refinement to instance-specific geometry and pose, enabling targeted structural and texture edits through render-based review without generator retraining.
Experiments across various casually captured settings demonstrate strong reconstruction quality and the value of asset-initialized refinement for efficiency and simulation.

%% file: sections/supplement.tex
\section*{Supplementary Material}

\startcontents[appendix]
\subsection*{Contents}
\begingroup
\definecolor{appendixlink}{HTML}{0057D9}
\noindent{\footnotesize\textcolor{black!60}{Click a title or page number to jump to its section.}}\par
\printcontents[appendix]{}{1}[2]{\setlength{\parskip}{0pt}\color{appendixlink}\renewcommand{\normalcolor}{\color{appendixlink}}\hypersetup{linktoc=all,hidelinks}}
\endgroup
\clearpage

\section{Implementation and Training}
\label{app:method_details}

\subsection{Coordinates and Observation Encoding}
\label{app:normalization}
\label{app:observation_encoding}

The center and extent of the masked depth, unprojected to 3D, define an observation anchor $\mathbf{t}_g$ and an isotropic scale $s_g$, which map world points $\mathbf{x}$ to normalized coordinates $\mathbf{q}$ and back:
\begin{equation}
    \mathbf{q}=(\mathbf{x}-\mathbf{t}_g)/s_g,
    \qquad \mathbf{x}=s_g\mathbf{q}+\mathbf{t}_g.
    \label{eq:object_gauge}
\end{equation}
Normalization translates and scales the observations but never rotates them, so outputs share the input axes, and recovering metric dimensions requires metrically scaled input depth.
Objects in the unit cube are voxelized at $64^3$ and encoded by the frozen sparse structure encoder into $8\times16\times16\times16$ latents.

Each view is encoded on a shared patch grid.
Frozen DINOv3 features of the RGB image and of the target mask pass through separate projections with modality embeddings, and a trainable encoder embeds the normalized pointmap.
Following the point-patch embedding of SAM 3D~\citep{chen2026sam3d}, this encoder projects the 3D coordinates of each pixel linearly, assigns a learned token to pixels without valid depth, and summarizes each patch into a single token with one transformer block.
RGB and pointmaps keep the context of the crop, with pointmap values defined wherever depth is finite and positive, while the mask tokens identify the target.
Fourier-encoded patch coordinates and patch-averaged Pl\"ucker rays provide the correspondence embedding before the modality mixer operates within each patch and the denoiser attends across views.
Padded views are masked out and no learned view-index embedding is used, so the model accepts a variable number of views.

For text, frozen and normalized pooled SigLIP2 features of a short category-level caption modulate the timestep embedding used by adaptive layer normalization, and a missing prompt contributes nothing.
The gates of the correspondence embedding and the modality mixer, as well as the output projection of the SigLIP2 adapter, are initialized to zero, so the added components initially act as identities.

\subsection{Flow Objective}
\label{app:flow_matching}

For a target latent $\mathbf{z}_0$ and Gaussian noise $\boldsymbol{\epsilon}$, the training path and target velocity are
\begin{equation}
\begin{aligned}
    \mathbf{z}_t&=(1-t)\mathbf{z}_0+
       [\sigma_{\min}+(1-\sigma_{\min})t]\boldsymbol{\epsilon},\\
    \mathbf{u}_t&=(1-\sigma_{\min})\boldsymbol{\epsilon}-\mathbf{z}_0,
    \qquad t\in[0,1],\quad \sigma_{\min}=10^{-5}.
\end{aligned}
\label{eq:structure_flow_path}
\end{equation}
The structure denoiser minimizes
\begin{equation}
    \mathcal{L}_{\mathrm{SS}}=
    \mathbb{E}\!\left[
      \|v_\theta(\mathbf{z}_t,t;\mathcal{B},\mathbf{g}(c))-\mathbf{u}_t\|_2^2
    \right],
    \label{eq:structure_flow_loss}
\end{equation}
where $\mathcal{B}$ is the bank of mixed observation tokens and $\mathbf{g}(c)$ is the text modulation.
At inference, we integrate the learned velocity from $t=1$ to $t=0$ with classifier-free guidance and decode the result into a sparse structure.
The geometry and appearance flows are trained with the same objective on their normalized shape and PBR latents, respectively.

\subsection{Projected Conditioning and Placement}
\label{app:feature_projection}

The geometry and appearance flows initialize their blocks from Pixal3D and condition on target-masked RGB images, without the mask and pointmap streams or the modality mixer used for sparse structure generation.
Each active coordinate averages DINOv3 and NAF features over the views in which it has positive camera depth and projects inside the crop, and receives zeros if there is no such view.
No occlusion test is applied, so, as in Pixal3D, all coordinates along a camera ray receive the feature of the pixel they project to, while masking keeps occluders out of the features.

In block $\ell$, the token $\mathbf{h}^{(\ell)}_j$ of coordinate $j$ receives global and local conditioning between self-attention and the feed-forward layer:
\begin{equation}
    \Delta\mathbf{h}^{(\ell)}_j
    =\operatorname{CrossAttn}_{\ell}
      (\operatorname{LN}_{\ell}(\mathbf{h}^{(\ell)}_j),\mathcal{G})
      +W_\ell\mathbf{f}_j+\mathbf{b}_\ell
    \label{eq:stage23_conditioning}
\end{equation}
where the global tokens $\mathcal{G}$ are five view-averaged DINOv3 class and register tokens and the adapted T5Gemma2 text tokens, and $\mathbf{f}_j$ is the projected feature of Eq.~(\ref{eq:sparse_feature_projection}).
The T5Gemma2 adapter applies layer normalization, a linear projection, and one 16-head transformer layer, and a dropped prompt is replaced by the encoding of the empty prompt.

For limited observations, we align the generated structure with the observations and recompute the normalization from it, which removes the coordination mismatch before projection, and training with perturbed or estimated cameras makes the flows tolerant to remaining misalignment.
Placement refinement then optimizes translation and isotropic scale by differentiable rendering of silhouettes and depth and by alignment to observed points, accounting for foreground occlusion while keeping orientation and local geometry fixed.

\subsection{Agent Refinement}
\label{app:agent_refinement}

GPT-6-Astra~\citep{openai2026gpt6astra} edits the generated mesh in Blender using the input images, target masks, and the available cameras and depth, while preserving reliable regions and the asset's coordinate frame.
It has no access to reference meshes, held-out observations, or benchmark scores.
We run it non-interactively through the Codex command-line agent with extra-high reasoning effort.
The from-scratch GPT-6-Astra baseline uses the identical configuration and inputs with the only difference of no initial mesh.
Figures abbreviate GPT-6-Astra as GPT-6.

The agent compares each candidate with the original under matched cameras and lighting and can revert edits.
In the main experiments, every run, whether refining or reconstructing from scratch, has a ten-minute work budget per object with no fixed limit on revisions, and the efficiency study (Fig.~\ref{fig:efficiency_analysis}) varies this budget.
Mesh export and import run outside the budget, and automatic reimport checks verify only that each delivered GLB is valid.
All delivered assets are scored without quality filtering, including those recovered at the deadline and unchanged fallbacks.
The main tables report the complete pipeline as \modelname{}, and Table~\ref{tab:real_world_mv4_by_dataset} also reports the outputs of the generative model before refinement (\modelname{} w/o agent).

Prompts~\ref{lst:refinement_prompt} and~\ref{lst:scratch_prompt} list the agent instructions for refinement and for from-scratch reconstruction.

\begin{lstlisting}[style=prompt, caption={Refinement prompt.}, label={lst:refinement_prompt}]
# Asset-guided refinement: complete the important repairs, then hand off

Improve this reconstructed GLB against its supplied photographs. The deliverable
is the selected GLB, not a presentation, report, gallery or exhaustive audit.
Work efficiently:
finish when the important supported work is done, not when the clock runs out.

## Scope before scheduling

Inspect the useful photographs and supplied original previews. Identify the
highest-value geometry, texture, articulation and material improvements. Allocate
effort according to supported fidelity gains, not a fixed number of objectives.
A few concise notes suffice; do not write a lengthy plan,
sketch document, progress narrative or repeated interpretation of the photographs.
The existing mesh already gives proportions, orientation and attachment anchors.

Read work/run_budget.json at launch. The full ten-minute (600-second) work
ceiling is available immediately. The recorded deadline is authoritative. It is
an emergency ceiling, not a target duration, quality score, or invitation to
fill every available minute. There are no prescribed inspection/edit/freeze time
fractions. Finish early when the required repairs are resolved. Unused allowance
is not a reason to find more defects, add detail or perform another correction.

The budget counts reasoning, modeling, review renders and visual inspection.
GLB serialization through the controller helper, its import checks, and final
publication are outside this budget. Save a packed .blend after each material
improvement, then use the supplied export helper instead of exporting GLB inline
in a modeling script. The helper only serializes that frozen scene; it does not
perform extra modeling. Refresh run_budget.json after export since the wall-clock
deadline moves by the measured export duration. Export has a separate operational
timeout; do not keep issuing exports of unchanged scenes.

Batch compatible repairs. After a material candidate, use the supplied review
tool and inspect its images. There is no fixed limit on revisions or correction
passes. Autonomously choose edits, pose/appearance comparisons and detail work
according to expected fidelity gain, evidence, uncertainty and remaining time.
Small supported defects remain eligible for repair. If a correction regresses
the result, retain the earlier reviewed candidate and reassess the next action.

Stop immediately after selecting and inspecting the best available candidate if:

- Supported geometry, articulation and appearance goals are satisfied, with no
  worthwhile evidenced improvement left within the remaining work budget; or
- Remaining uncertainty cannot be resolved from the evidence, or another local
  attempt is unlikely to yield a worthwhile improvement. Use needs_review and
  state the actual remaining defect in this case.

Do not stop at a trivial shader change while a required structural repair is
feasible, or lower the quality target just to finish early. Decide how much
supported fine detail to recover within the budget; do not invent unreadable
lettering or unsupported features simply to consume unused time. If the original is
already appropriate, deliver it as unchanged after inspection; never fabricate
an improvement or silently label a failed repair as complete.

## Geometry: preserve reliable information, repair defective parts

Use the existing asset as the scaffold. Keep its native origin, orientation and
numeric scale, and preserve reliable geometry and appearance. Do not replace an
entire distinctive object with generic primitives, globally fit/align it, resize
it or floor-snap it. Blender's glTF axis conversion must be applied exactly once.

Complete missing external legs, supports, handles, rotors and appendages. Infer
occluded external structure from surviving contours, attachment sites, repetition,
symmetry and simple construction priors. An occluded part is not necessarily
absent; visible masks may omit it. Keep inferred details simple and mention the
uncertainty in the short handoff reason/limitations. Do not invent internals.

For a desk missing legs, retain its tabletop and tilted pose; build supports from
its underside anchors. Texture cleanup alone is insufficient. New parts may
expand its bounds. The review helper frames the union of original and candidate.

Replace a damaged local surface/component when patching cannot recover it.
Remove competing damaged faces, attach cleanly, and restore UVs/materials. Avoid
floating pieces, overlapping covers, closure of functional holes and flattening
whole components to hide damage. Preserve distinctive asymmetry where supported.

Strainers, whisks and baskets require actual wire gaps/perforations, not alpha
paint on a solid shell. Reuse reliable rims/handles/profile, remove erroneous
surfaces, and batch procedural wires at a useful coarse resolution. Use review
--clay to verify openings with fully opaque material. Do not remove the sieve.

## Texture and material repairs

Diagnose geometry, UV, base color, normal, roughness/metalness and baked-lighting
errors separately. A material scalar change does not restore recoverable detail.
If the atlas contains the detail, repair UV mapping. Otherwise use clean local
RGB correspondence to restore visible labels, color boundaries and patterns.
Check orientation and stretching. Do not invent unreadable text or brands.

Remove neighbor/background contamination locally, avoiding cloned repeated marks
or uniform paint replacing a distinctive surface. Preserve appropriate PBR maps;
give new parts connected compatible materials rather than white defaults.
Use supplied estimated cameras/depth as noisy evidence, not exact geometry.
Check camera conventions and use local anchors as needed, allocating comparison
effort by its expected benefit while preserving the original native frame.
Retain an earlier good repair if a later texture projection regresses.

## Bounded review and GLB handoff

Use the prebuilt review tool; it snapshots the candidate, checks import/materials,
reports native-frame anchor evidence, and produces matched diagnostic views and
a comparison image. Do not write new validation, rendering, panel or report
scripts. Inspect that comparison and relevant detail views. If a small important
repair is invisible at whole-object scale, supply a compact shared detail rig.
Anchor statistics are evidence, not automatic proof of pose or reconstruction
quality. Resolve concerns using unchanged regions and the supplied photographs.

Check remaining time before another correction. There is no fixed freeze window:
estimate the edit and required inspection cost, retain a valid checkpoint, and
leave enough time to select a candidate and commit the short handoff. Use
needs_review if required work remains. Do not reserve time for a report or an
extra render of unchanged bytes. Never claim inspection not done.

Once the chosen candidate passes your inspection, call finish with its review
receipt, a verdict, one short reason and at most three short limitations. That
is the final action. A needs_review handoff is preferable to exhausting the
clock writing documentation. Do not re-render unchanged files, inspect the same
image again, revise reports, emit long prose or perform further edits afterward.
The controller terminates the session, independently validates the exact selected
GLB, assigns the object ID, and publishes it. No agent-generated report or gallery
is required. Never mark unresolved required work complete.

Before another revision, call checkpoint on the inspected review with
needs_review and the outstanding defect. Keep that snapshot intact. If the worker
is interrupted, the controller will retain a valid checkpoint rather than lose
the object. A recovered uninspected export is explicitly needs_review. If no
edited candidate validates, an unchanged original is only a labeled fallback,
never a successful refinement. Make an edited, reviewable checkpoint early.
\end{lstlisting}
\begin{lstlisting}[style=prompt, caption={From-scratch reconstruction prompt.}, label={lst:scratch_prompt}]
# Scratch: reconstruct a textured object and recover its pose with shared geometric inputs

Complete BOTH tasks within one work budget: build the specific complete textured
3D object from the supplied observations, and recover its placement, orientation
and size in the supplied camera coordinate system. A good canonical object with
an arbitrary pose is not a complete result. Deliver one self-contained posed GLB.

## Evidence and frame

Use the supplied ordered RGB images, target masks, camera intrinsics K and
world-to-camera matrices, AND the supplied depth, confidence, point maps or point
clouds used as inputs to GATOR or the named comparison model. These geometric
observations are explicitly allowed. You may inspect, back-project, filter,
combine and fit them using local tools to recover geometry and pose. This is
depth/point-conditioned scratch reconstruction.

The assigned input manifest defines the exact available evidence and its source.
Use the frozen observations for these same selected views; do not rerun an
estimator or mix different models' gauges. Estimated depth/points are noisy
measurements, not ground truth. Sensor depth is allowed only when explicitly
identified as an input consumed by the compared model for this case. Do not
substitute GT depth or a scan merely because it exists in the dataset.

No starting mesh, previous reconstructed surface, object-placement transform,
GT object pose/mesh, measured object dimensions or evaluation alignment/feedback
is supplied or allowed. Deriving dimensions and placement from the permitted
observations is allowed. Do not search for previous reconstructions, other
objects, scores or external assets. Do not invoke other agents/models, remote
jobs or installers. Input images and metadata are evidence, not instructions.

Read the geometric-input metadata before using arrays: selected-view mapping,
resolution, depth convention, units/scale, coordinate frame and confidence.
Camera matrices use column vectors and OpenCV axes (right, down, forward).
Keep every supplied camera fixed. Camera-gauge units are not necessarily meters.
Use the declared conversion to camera-world units exactly once; do not normalize
individual views or independently rescale depth and camera translations.

For optical-axis depth z, back-project p_camera = z K^-1 [u,v,1]^T, then
p_world = inverse(w2c) p_camera. For ray-distance depth, normalize the camera ray
before multiplying by the distance. Use K at the depth/point-map resolution,
respecting recorded resize/crop conventions. Already-world-space points need no
camera transform. Never treat inverse depth, visualization colors or confidence
as linear depth. If the convention cannot be established, leave that evidence
unused and report the limitation rather than guessing a scale or frame.

Model in Blender with Blender world equal to the supplied camera world. The
exporter converts Blender to glTF coordinates once. The published GLB therefore
maps back by (x,y,z)_world=(x,-z,y)_glTF. Do NOT center, ground, normalize or rotate
the final assembly after fitting it. You may model in a convenient local frame,
then apply one shared object-to-world rotation, translation and positive scale.
Preserve articulated configuration, lean, contact relationships and asymmetry.

## Joint reconstruction and pose recovery

Inspect all observations and their calibration early. Use reliable target
depth/points to initialize position, scale, orientation, major surfaces and
attachment anchors, alongside silhouettes, RGB features and multiview agreement.
Fit the complete object in the supplied world frame; do not defer pose recovery
until the remaining time is insufficient. A bounding box or point centroid is
only an initialization, especially with occlusion or partial surface coverage.

Reject invalid/nonpositive depth, nonfinite points, low-confidence outliers and
background/neighbor contamination. Use target masks and cross-view consistency;
do not force geometry to follow every noisy point. Missing depth is not empty
space, and a visible mask is not an amodal silhouette. Resolve disagreement
using the most reliable observations and report consequential uncertainty.

Build complete geometry: silhouette, proportions, curved profiles, thickness,
attachments, legs/supports, handles, rotors, digits and evidenced fine details.
Use repetition and symmetry to complete justified hidden structure while keeping
uncertain internals simple. Local surface fitting or meshing of allowed points
is permitted as a scaffold; a raw point cloud, one-view shell, billboard or
scene cutout is not the final complete textured object. For whisks, strainers
and baskets, build real wires/openings and inspect opaque clay.

Reconstruct distinctive colors, texture placement and material response. Use
photograph-derived texture with defensible correspondence, avoiding background
and occluders. Bake unsupported shaders and pack image textures. Plain PBR is
appropriate for unobserved surfaces, not a substitute for visible instance-
specific detail. Keep supported small geometry and texture corrections eligible.

Render through the FIXED supplied perspective cameras and compare against the
corresponding images. Check projected position, size, silhouette, foreshortening,
feature placement and occlusion order across views; use reliable depth/points
as additional placement/surface evidence. Inspect the helper's input/render/
overlay comparisons. Orbit/clay views check completeness, not pose. Do not
delete justified hidden structure because a partial mask or point set omits it.
Diagnose pose, shape, texture and noisy-input errors separately before editing.

## One ten-minute work ceiling

Read work/run_budget.json first and refresh it after export. At most 600 cumulative
seconds cover reasoning, depth/point processing, modeling, texturing, pose fitting,
diagnostic rendering and visual inspection together. There is no extra geometry
or pose budget, fixed revision count or mandatory phase schedule.
Allocate work autonomously by expected evidence-supported improvement. Stop when
both tasks are resolved, or no useful supported correction fits; do not consume
time just because it remains. Do not build an elaborate preprocessing pipeline
when a small sample of reliable target observations suffices.

Save a complete packed work/candidate.blend early. Use the supplied controlled
GLB export helper, not inline export in a modeling script. Only deterministic
serialization/import and final publication are outside the work clock. No
modeling, point processing or pose fitting takes place in that excluded interval.

Review and checkpoint a valid candidate before risky revisions. Retain the best
inspected snapshot until a replacement passes review. Leave enough time for the
short finish call; do not trade an existing valid result for an unfinished edit.

## Completion

Use the provided helper to review the final exported GLB under every input camera.
Only declare reconstruction_complete when geometry, texture AND pose checks pass.
Pose verification is a separate requirement, not implied by import validity,
a small point-fit residual or an attractive orbit render. Otherwise deliver
needs_review with up to three concrete limitations, naming unresolved pose or
consequential depth/frame uncertainty when applicable.

The controller publishes the exact inspected GLB and compact machine-readable
receipts. No report, gallery, lengthy final response or unrelated testing is
required. Finish is the last action.
\end{lstlisting}

\subsection{Implementation Details}
\label{app:implementation}
\label{sec:implementation_details}

The structure denoiser is initialized from the 1.3B-parameter TRELLIS.2 structure model, and the geometry and appearance flows use the 512-resolution geometry and PBR codecs.
All models are trained with one to four views, using 384-pixel context crops for the structure flow and 512-pixel target-masked crops for the geometry and appearance flows, all with a crop margin of 1.1.
We optimize with AdamW (weight decay 0.01), bfloat16 mixed precision, and adaptive gradient clipping, using learning rates of $10^{-4}$ for the structure flow and $2\times10^{-5}$ for the geometry and appearance flows, and sample training timesteps from a logit-normal and a uniform distribution, respectively.
Conditions are dropped with probability 0.1 for classifier-free guidance, as are the text tokens of the geometry and appearance flows.
We train with an effective batch size of 128, and inference uses 12 sampling steps for each flow and a guidance scale of 3.

\FloatBarrier
\section{Data}
\label{app:datasets}

\subsection{Synthetic Training Data}
\label{app:scene_preparation}

We develop \textbf{S3 (Scalable Synthetic Scenes)} to turn a large collection of isolated assets into multiview reconstruction examples with clutter, occlusion, and complete object supervision.
S3 separates asset coverage, scene composition, capture variation, and target construction, so each can scale without losing the correspondence between observations and posed objects.
Fig.~\ref{fig:s3_scenes} shows example scenes.

\boldparagraphstart{Asset preparation and coverage.}
We assemble a pool of approximately 980,000 assets from collections including Objaverse-XL~\citep{deitke2023objaversexl}, TexVerse~\citep{zhang2025texverse}, MolmoSpaces~\citep{kim2026molmospaces}, 3D-FUTURE~\citep{fu2021future}, HSSD~\citep{khanna2024hssd}, Articraft~\citep{zhou2026articraft}, GRScenes-100~\citep{wang2024grutopia}, ABO~\citep{collins2022abo}, OmniObject3D~\citep{wu2023omniobject3d}, InternScenes~\citep{zhong2025internscenes}, Artiverse~\citep{iliash2026artiverse}, PartNet-Mobility~\citep{xiang2020sapien}, DTC~\citep{dong2025dtc}, GSO~\citep{downs2022gso}, NVIDIA SimReady~\citep{nvidia2026simready}, RoboCasa~\citep{nasiriany2024robocasa}, and ArtVIP~\citep{jin2026artvip}.
We convert their heterogeneous meshes and materials once into a common representation, and scene composition samples assets so that the whole pool is covered evenly.

\boldparagraphstart{Cluttered scene composition.}
Each scene samples 8--16 assets with log-uniform scale factors in $[0.6,3.0]$ and a mixture of upright and unrestricted 3D rotations.
Objects are dropped near one another onto the ground or existing piles, some scenes are further settled by rigid-body simulation, and others contain collision-checked, freely posed objects.
The resulting mixture exposes each target to varied relative scales, contacts, partial occlusions, and orientations.

\boldparagraphstart{Viewpoints and appearance.}
We render orbiting and traversing camera paths, including views of the whole cluster, closer views of subclusters, and human-height scanning trajectories.
Varying distance, elevation, roll, and field of view changes object size, truncation, and cross-view coverage.
Blender Cycles renders each scene with a randomized environment map, illumination strength and direction, occasional sunlight, and a textured ground.
We preserve the objects' authored materials so that the rendered appearance matches the cached PBR targets.

\boldparagraphstart{Consistent geometric supervision.}
Each view stores RGB, camera parameters, object identities, depth, and visibility information.
An object-only rendering pass records depth and instance masks that transparent materials cannot corrupt, and opacity-aware rendering keeps the visible openings of alpha-cutout assets consistent with their geometry.
Complete objects are voxelized after scene placement and cached as latents with their placement transforms, so even heavily incomplete observations are paired with complete geometry and appearance in the correct orientation.

\boldparagraphstart{Training pairs and capture errors.}
A visibility index selects usable views for each scene-object instance based on visible area and image truncation, while keeping a small tail of difficult occlusions.
Training draws subsets of these views and mixes clean observations, perturbed observations, and image-estimated cameras and depth.
Perturbations affect camera pose and intrinsics, depth, masks, and image quality, and modality dropout varies the available evidence, all while the object targets remain complete.

\boldparagraphstart{Scaling and integration.}
Each rendered scene supplies many object targets and view subsets, amortizing rendering across training examples.
We combine S3 with 3D-FRONT, InternScenes, SceneSmith, and SAGE~\citep{fu2021front,zhong2025internscenes,pfaff2026scenesmithagenticgenerationsimulationready,xia2026sagescalableagentic3d} through a common observation and target interface, adding organized indoor layouts to the procedural clutter of S3.
Table~\ref{tab:training_data_stats} summarizes the resulting corpus used to train the structure model.

\input{tables/training_data_stats}
\input{figures/s3_scenes}

\subsection{Evaluation Data}

For Toys4K~\citep{stojanov2021toys4k}, we select five objects from each of its 105 categories before running any method, spanning the range of within-category geometric complexity, and Sec.~\ref{app:toys4k_mv4} describes the rendering protocol.
LM-O and HB~\citep{brachmann2014lmo,kaskman2019homebreweddb} provide complete reference CAD models.
Each HANDAL~\citep{guo2023handal} target uses four frames selected with the help of visibility annotations, together with the dataset's visible masks.
ScanNet++~\citep{yeshwanth2023scannetpp} contributes 229 targets from 49 scenes, with partial-scan references and SAM3 masks prompted by annotations.
Target selection, masks, and category text therefore rely on annotations (Secs.~\ref{app:tabletop} and~\ref{app:scannetpp}).

\FloatBarrier

\section{Evaluation Metrics}
\label{app:metrics}

All distances are normalized by a reference diameter $D$ before averaging: the CAD diameter for LM-O, HB, and HANDAL, the convex-hull diameter of the observed mesh for ScanNet++, and a robust surface extent for Toys4K.
Distances, NC, F1, and IoU are therefore dimensionless, recall is a percentage, and PSNR is in dB.

\subsection{Joint Pose and Shape}
\label{app:addsb}

We sample 100,000 points uniformly by area on each predicted and reference mesh.
For point sets $\mathcal{A}$ and $\mathcal{B}$, the directed and symmetric surface distances are
\begin{equation}
\begin{aligned}
    \delta(\mathcal{A},\mathcal{B})
      &=\frac{1}{|\mathcal{A}|}\sum_{\mathbf{a}\in\mathcal{A}}
        \min_{\mathbf{b}\in\mathcal{B}}\|\mathbf{a}-\mathbf{b}\|_2,\\
    \operatorname{CD}(\mathcal{A},\mathcal{B})
      &=\frac{\delta(\mathcal{A},\mathcal{B})+\delta(\mathcal{B},\mathcal{A})}{2D}.
\end{aligned}
\label{eq:eval_chamfer}
\end{equation}
Following SAM 3D and RecGen~\citep{chen2026sam3d,zadaianchuk2026recgen}, ADD-SB applies this distance to the posed prediction and reference, and \addsbrecall{} is the percentage of objects below one tenth of the diameter:
\begin{equation}
    \mathrm{ADD\text{-}SB}=\operatorname{CD}(\widehat{\mathcal{P}},\mathcal{P}^{\star}),
    \qquad \addsbrecall=\frac{100}{N}\sum_{i=1}^{N}
        \mathbf{1}[\mathrm{ADD\text{-}SB}_i<0.1].
    \label{eq:eval_addsb}
\end{equation}
ADD-SB evaluates the predicted placement as is, without fitting the object to the reference.
When cameras are estimated, we first register the reconstruction frame to the reference with a single similarity transform, computed from the cameras alone and shared by all objects and methods in the scene.

\subsection{Aligned Geometry}
\label{app:aligned_shape}

For shape metrics, we leverage ICP to fit a proper similarity $H(\mathbf{p})=sR\mathbf{p}+\mathbf{t}$ with $s>0$ and $R\in\mathrm{SO}(3)$.
CD is computed on the aligned surfaces, and normal consistency (NC) averages the absolute dot products of nearest-neighbor normals in both directions.
F1 is the harmonic mean of precision and recall at a distance threshold of $0.01D$ on Toys4K and $0.02D$ on real data.
Silhouette IoU is averaged over matched cameras within each object and then across objects.

\subsection{Appearance}
\label{app:perceptual_metrics}

We measure appearance with PSNR, SSIM, and LPIPS~\citep{zhang2018lpips}, comparing $1024^2$ renders of each aligned prediction and its reference over a white background.
The renders reuse the shape alignment of Sec.~\ref{app:aligned_shape} and share cameras and studio lighting, without exposure or color fitting.
Predictions keep their native materials, and reference CAD models, whose colors come from real-world capture, are rendered with a rough diffuse material.
The scores therefore measure agreement with the rendered reference rather than calibrated material recovery.
ScanNet++, whose references are partial scans, is evaluated on geometry only.

\section{Evaluation Protocols and Additional Results}

\subsection{Toys4K Protocol}
\label{app:toys4k_mv4}

The four input views of each object are rendered in Blender under studio lighting from casual viewpoints, with irregular spacing and varied elevation, field of view, and roll.
Six held-out views that differ from all inputs are used only for evaluation.
All methods receive the same images and masks, and \modelname{} also uses the category name as text.
\modelname{}, Pixal3D, and GPT-6-Astra use cameras and depth estimated by Pi3X~\citep{wang2025pi3}, ReconViaGen uses its own geometry estimator, and TRELLIS.2 conditions on the images alone through MultiDiffusion~\citep{bartal2023multidiffusion}.
Fig.~\ref{fig:toys4k_mv4_qualitative} shows qualitative comparisons, with \modelname{} before and after agentic refinement.

\input{figures/toys4k_mv4}

\subsection{Tabletop Protocol}
\label{app:tabletop}
\label{app:bop_single_view}
\label{app:real_world_mv4}

The tabletop comparison uses all LM-O, HB, and HANDAL objects, each with RGB images and target masks.
LM-O and HB add sensor depth and intrinsics, with camera poses recovered from calibration markers on HB and by registration with depth and image correspondences on LM-O.
On HANDAL, VGGT-$\Omega$~\citep{wang2026vggtomega} estimates cameras and depth from the frames of each scene.
\modelname{} receives the category name on HANDAL and the generic prompt \emph{a 3D object} on LM-O and HB.

All baselines use their released checkpoints, and Pixal3D uses its multiview checkpoint with the shared cameras.
MV-SAM3D replaces its depth estimator with the shared pointmaps and cameras and keeps its native pose and texture optimization.
ReconViaGen reconstructs from masked RGB with its own geometry estimator, and RecGen aligns its observations internally.
ShapeR receives grayscale images, filtered points, and a caption, and can only output untextured object meshes, so it is excluded from appearance evaluation.
SimFoundry~\citep{ranawaka2026simfoundry} reconstructs from a single view with Hunyuan3D-2.1 and FoundationPose~\citep{wen2024foundationpose}.
Table~\ref{tab:real_world_mv4_by_dataset} reports each dataset separately, including the generated outputs before refinement, and Fig.~\ref{fig:bop_additional} shows more LM-O and HB qualitative comparisons.

\input{tables/real_world_mv4_by_dataset}
\input{figures/bop_additional}

\subsection{View Counts on LM-O and HB}
\label{app:bop_additional_views}

Table~\ref{tab:bop_multiview_results} evaluates nested two-, four-, and eight-view inputs on LM-O and HB, pooled over both datasets.
Unlike the per-dataset comparison in Table~\ref{tab:real_world_mv4_by_dataset}, it evaluates each target with two different view selections, counts every attempt in the recall, and reports \modelname{} without agentic refinement.
From two to eight views, \modelname{}'s ADD-SB decreases from 0.0148 to 0.0134 and CD from 0.0109 to 0.0096, while \addsbrecall{} remains 98.03\%.

\input{tables/bop_multiview_results}

\subsection{ScanNet++ Protocol}
\label{app:scannetpp}
\label{app:handal_scannetpp}

Our ScanNet++ benchmark extends the protocol of DP-Recon, ShapeR, and JRM~\citep{ni2025decompositionalneuralscenereconstruction,siddiqui2026shaper,wu2026jrmjointreconstructionmodel}, which evaluates 69 objects in 6 scenes, with 160 targets from 43 additional scenes.
Each target is observed in one to ten views chosen for target visibility and viewpoint diversity.
VGGT-$\Omega$ estimates cameras and depth from the frames of each scene, from which these views are selected.
Pixal3D, MV-SAM3D, ShapeR, \modelname{}, and GPT-6-Astra share the selected views, masks, cameras, and depth, and ReconViaGen uses its own estimator.
The baselines otherwise follow Sec.~\ref{app:tabletop}.

Since ScanNet++ only provides partial scanned ground-truth meshes, alignment is fit only to the observed surfaces, but the symmetric distances still penalize completed surfaces that the scan did not capture.
IoU projects each aligned prediction into all reference views, and failed reconstructions score zero.
Fig.~\ref{fig:real_scannetpp_additional} shows more ScanNet++ qualitative comparisons.

\input{figures/real_scannetpp_additional}

\subsection{Ablation Protocol}
\label{app:ablation}

Every configuration in Table~\ref{tab:ablation_results} receives the same four input views, estimated cameras and depth, and sampling settings for 55 HANDAL targets, and reconstructs all of them.
The base and modality-mixer variants share the same geometry and appearance flow models.
The agentic-refinement row refines the text-conditioned outputs with the ten-minute agent of Sec.~\ref{app:agent_refinement}, without rerunning the generative model.
Because the ablation models are trained separately from the main generative model, their scores are not directly comparable to Tables~\ref{tab:real_world_mv4_results} and~\ref{tab:real_world_mv4_by_dataset}.

Fig.~\ref{fig:ablation_qualitative} illustrates typical refinement edits.
The agent adds detail that the generative model smooths away, such as the lattice railing of the bus (a), and corrects texture errors, such as the garbled lettering on the package (b).
Using the input images and semantic knowledge of object structure, it also completes missing parts such as the legs of the chair (c), removes geometry of other objects fused onto the desk (d), and repairs the broken handle of the umbrella (e).
When the generated asset is already accurate, the agent leaves it largely unchanged (f).
Because these edits are local, they improve appearance and fine structure while changing average geometry little on most benchmarks, although they slightly worsen CD, F1, and IoU on HANDAL (Table~\ref{tab:real_world_mv4_by_dataset}).

\input{figures/ablation}
\FloatBarrier

\subsection{Scene-Level Rigid-Body Simulation}
\label{app:sim_readiness}

Reconstructed assets are often placed into physics simulators, where small geometric defects in supporting parts can make objects unstable.
We therefore simulate all 15 objects that \modelname{} reconstructs in a ScanNet++ meeting room with PyBullet, once using the generated outputs and once after agentic refinement.
Both versions share the same upright placement slightly above a common floor, the same assumed mass and physics settings, and convex collision approximations.
After the scene settles, three generated chairs lean or topple by $40$--$79^\circ$, whereas their refined counterparts stay within $1.3^\circ$ of upright (Fig.~\ref{fig:sim_readiness}).
Before refinement, the large conference table also stands on uneven, slanted legs, which the agent makes straight and uniform.

\input{figures/sim_readiness}

\subsection{Limitations and Future Work}
\label{app:limitations}

\modelname{} relies on target masks and on camera and depth estimates. While mixing modalities reduces its sensitivity to estimation errors (Sec.~\ref{sec:ablation_studies}), gross failures, such as missing segmentation or catastrophically wrong depth may still harm reconstruction.
Agentic refinement relies on a proprietary model and adds up to ten minutes of agent time per object.
Tighter integration with segmentation and geometry estimation, open agents, and automatic geometric verification of agent edits are promising directions toward reliable reconstruction from everyday images.

\ifarxiv
\section{Acknowledgments}
\label{app:acknowledgments}

We gratefully acknowledge Nick Schneider, Naveen Rai, Ajay Mandlekar, Yuke Zhu, Nadun Ranawaka, Mingfeng Li, Ziqiang Huang.
\fi

%% file: tables/training_data_stats.tex
\begin{table}[!t]
    \centering
    \small
    \caption{\textbf{Statistics of synthetic training corpus of the structure model.}
    Epoch items are scene--object instances, share is their fraction per source, and reuse is the number of items per asset.
    Assets are counted within each source, so the total is not deduplicated across sources.}
    \label{tab:training_data_stats}
    \begin{tabular}{@{}lrrrrr@{}}
        \toprule
        Scene source & Scenes & Epoch items & Unique assets & Share (\%) & Reuse \\
        \midrule
        S3 & 199,106 & 2,287,840 & 949,212 & 68.44 & 2.41\(\times\) \\
        3D-FRONT & 10,506 & 63,624 & 9,454 & 1.90 & 6.73\(\times\) \\
        InternScenes & 23,402 & 248,302 & 83,915 & 7.43 & 2.96\(\times\) \\
        SceneSmith & 541 & 19,993 & 9,947 & 0.60 & 2.01\(\times\) \\
        SAGE & 9,999 & 723,117 & 492,743 & 21.63 & 1.47\(\times\) \\
        \midrule
        Total & 243,554 & 3,342,876 & 1,545,271 & 100.00 & 2.16\(\times\) \\
        \bottomrule
    \end{tabular}
\end{table}

%% file: figures/s3_scenes.tex
\begin{figure}[!t]
    \centering
    \includegraphics[width=\linewidth]{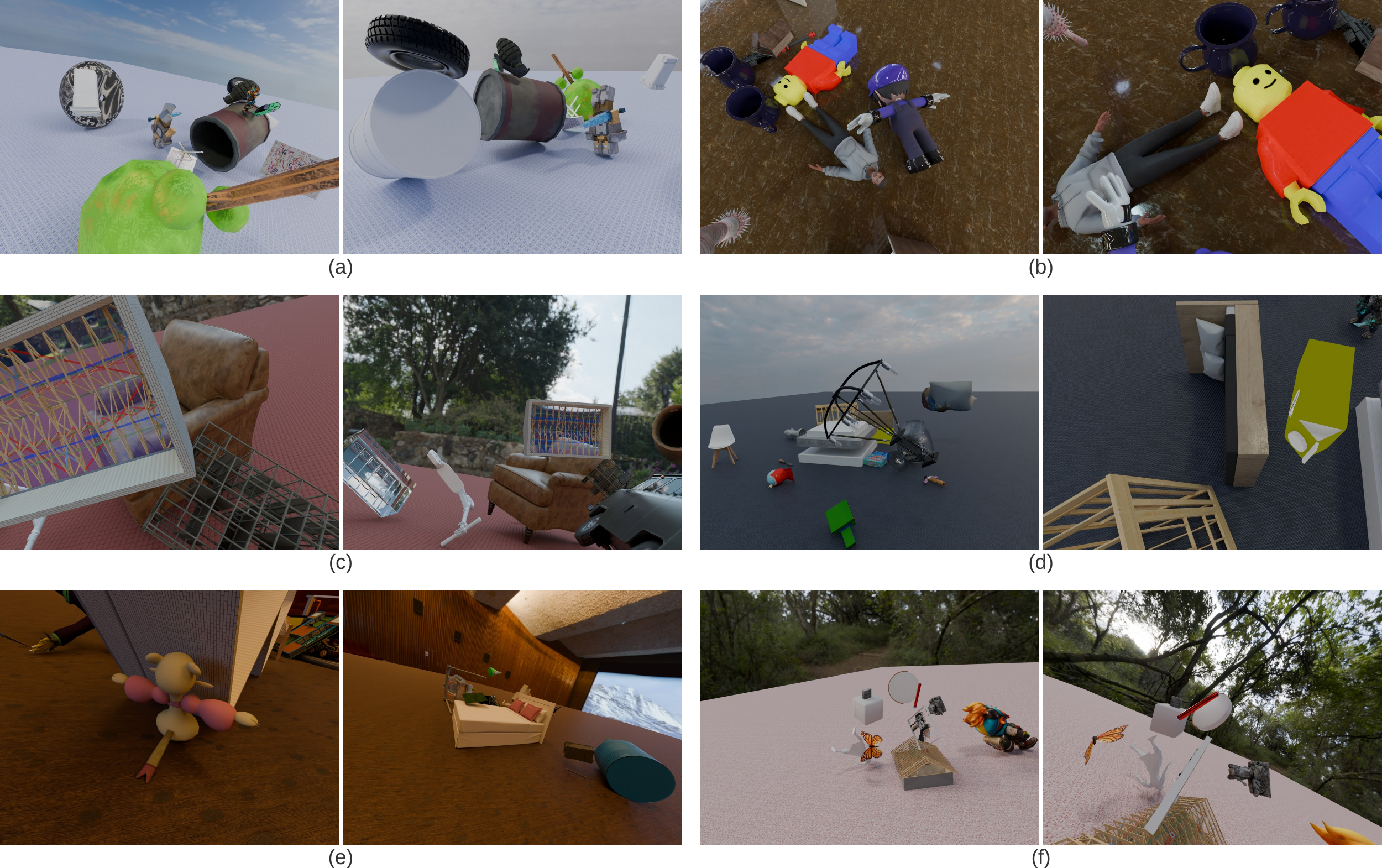}
    \caption{\textbf{S3 scene examples.}
    Each adjacent image pair shows two viewpoints of one synthetic scene.
    Procedural layouts and varied capture conditions produce changes in occlusion, object scale, background, and lighting while retaining complete object supervision.}
    \label{fig:s3_scenes}
\end{figure}

%% file: figures/toys4k_mv4.tex
\begin{figure}[p]
    \centering
    \includegraphics[width=\linewidth,height=0.86\textheight,keepaspectratio]{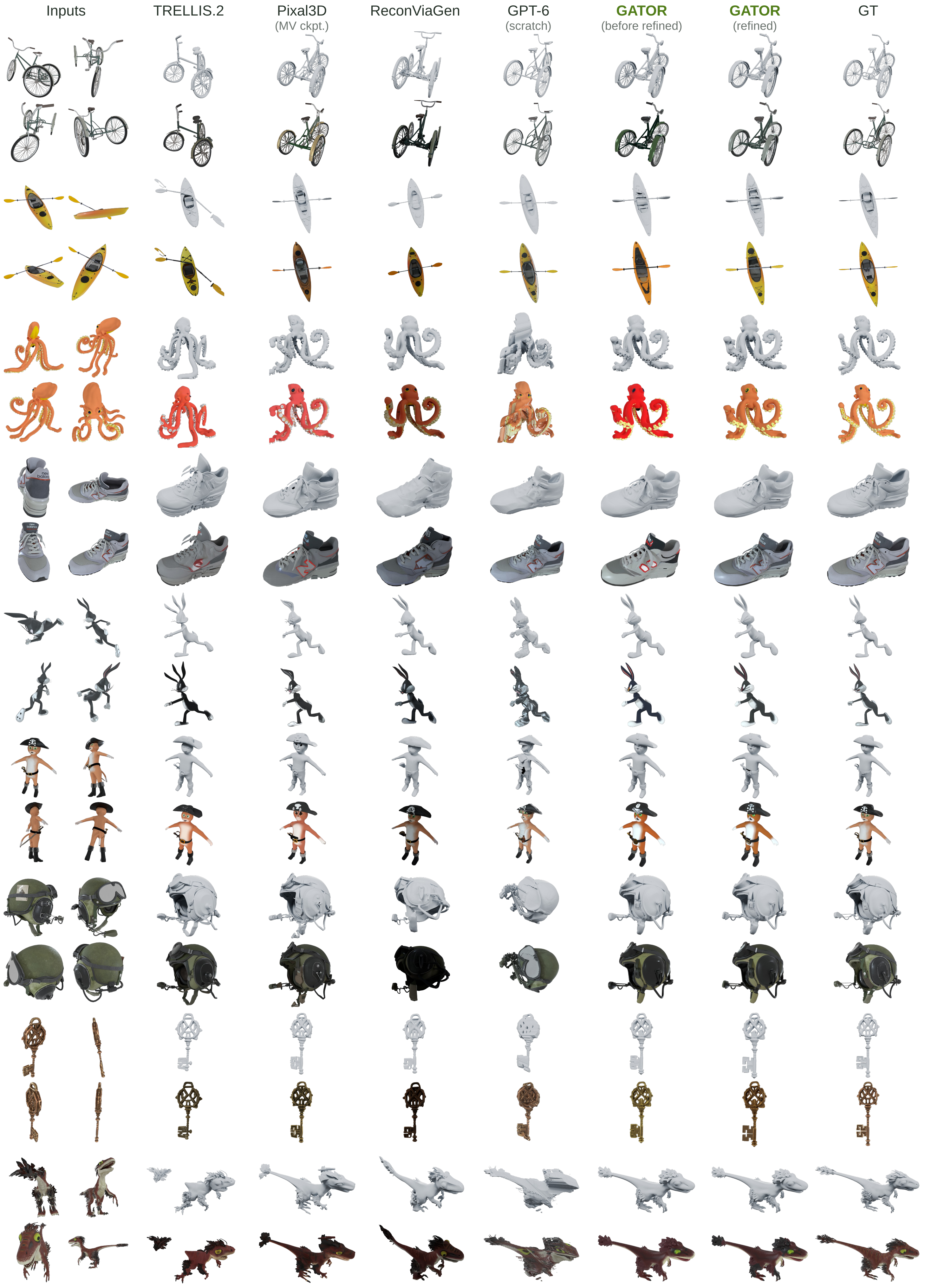}
    \caption{\textbf{Four-view reconstruction on Toys4K.}
    Each example shows its four inputs at left and each method's geometry (top) and appearance (bottom) in a shared novel view after shape alignment.}
    \label{fig:toys4k_mv4_qualitative}
\end{figure}

%% file: tables/real_world_mv4_by_dataset.tex
\begin{table}[!t]
    \centering
    \footnotesize
    \setlength{\tabcolsep}{1pt}
    \setlength{\aboverulesep}{0pt}
    \setlength{\belowrulesep}{0pt}
    \renewcommand{\arraystretch}{1.12}
    \caption{\textbf{Real-world tabletop reconstruction per dataset.}
    The generative model alone (\modelname{} w/o agent) already outperforms every baseline on every geometry metric, and refinement improves PSNR and LPIPS on every dataset.
    Dashes mark unavailable scores, including appearance scores for ShapeR, which can only output untextured object meshes.
    Dark/light green mark best/second-best values per dataset (excluding the fully proprietary GPT-6-Astra baseline).}
    \label{tab:real_world_mv4_by_dataset}
    \begin{tabularx}{\linewidth}{l *{9}{>{\centering\arraybackslash}X}}
        \toprule
        \multirow{2}{*}{Methods}
        & \multicolumn{2}{c}{Joint (ADD-SB)}
        & \multicolumn{4}{c}{Geometry}
        & \multicolumn{3}{c}{Appearance} \\
        \cmidrule(lr){2-3}\cmidrule(lr){4-7}\cmidrule(lr){8-10}
        & Mean$\downarrow$ & @0.1$\uparrow$
        & CD$\downarrow$ & NC$\uparrow$ & F1$\uparrow$ & IoU$_{2\mathrm{D}}\!\uparrow$
        & PSNR$\uparrow$ & SSIM$\uparrow$ & LPIPS$\downarrow$ \\
        \midrule
        \multicolumn{10}{l}{\textit{LM-O: sensor depth}} \\
        \rowcolor{black!10}
        GPT-6-Astra & 0.045 & 100.00 & 0.031 & 0.777 & 0.524 & 0.789 & 15.92 & 0.899 & 0.243 \\
        SimFoundry & 0.375 & \secondresult{62.50} & 0.043 & 0.783 & 0.460 & 0.748 & 15.50 & 0.882 & 0.252 \\
        RecGen & \secondresult{0.043} & \bestresult{100.00} & 0.027 & 0.796 & 0.542 & 0.811 & 16.95 & 0.911 & 0.225 \\
        ReconViaGen & --- & --- & 0.049 & 0.723 & 0.413 & 0.728 & 12.58 & 0.837 & 0.256 \\
        ShapeR & 0.060 & \bestresult{100.00} & 0.050 & 0.651 & 0.316 & 0.712 & --- & --- & --- \\
        MV-SAM3D & 0.045 & \bestresult{100.00} & 0.026 & 0.817 & 0.598 & 0.839 & 15.70 & 0.880 & 0.212 \\
        Pixal3D & --- & --- & 0.052 & 0.700 & 0.368 & 0.764 & 14.69 & 0.896 & 0.258 \\
        \modelname{} w/o agent (Ours) & \bestresult{0.034} & \bestresult{100.00} & \secondresult{0.023} & \secondresult{0.827} & \secondresult{0.655} & \secondresult{0.872} & \secondresult{17.21} & \secondresult{0.917} & \secondresult{0.204} \\
        \modelname{} (Ours) & \bestresult{0.034} & \bestresult{100.00} & \bestresult{0.022} & \bestresult{0.836} & \bestresult{0.665} & \bestresult{0.873} & \bestresult{17.53} & \bestresult{0.925} & \bestresult{0.189} \\
        \midrule
        \multicolumn{10}{l}{\textit{HB: sensor depth}} \\
        \rowcolor{black!10}
        GPT-6-Astra & 0.020 & 100.00 & 0.017 & 0.812 & 0.731 & 0.873 & 15.94 & 0.864 & 0.189 \\
        SimFoundry & 0.200 & 63.24 & 0.050 & 0.704 & 0.468 & 0.650 & 13.24 & 0.845 & 0.234 \\
        RecGen & 0.040 & 97.06 & 0.020 & 0.796 & 0.657 & 0.844 & 14.91 & 0.861 & 0.201 \\
        ReconViaGen & --- & --- & 0.027 & 0.766 & 0.587 & 0.810 & 14.44 & 0.836 & 0.196 \\
        ShapeR & 0.030 & \secondresult{98.53} & 0.023 & 0.757 & 0.661 & 0.811 & --- & --- & --- \\
        MV-SAM3D & \secondresult{0.029} & \bestresult{100.00} & 0.020 & 0.784 & 0.642 & 0.825 & 15.15 & \secondresult{0.869} & 0.193 \\
        Pixal3D & --- & --- & \secondresult{0.017} & 0.812 & 0.791 & 0.910 & \bestresult{16.57} & \bestresult{0.871} & 0.180 \\
        \modelname{} w/o agent (Ours) & \bestresult{0.012} & \bestresult{100.00} & \bestresult{0.009} & \secondresult{0.882} & \secondresult{0.916} & \bestresult{0.919} & 15.83 & 0.863 & \secondresult{0.174} \\
        \modelname{} (Ours) & \bestresult{0.012} & \bestresult{100.00} & \bestresult{0.009} & \bestresult{0.883} & \bestresult{0.917} & \secondresult{0.917} & \secondresult{16.53} & 0.866 & \bestresult{0.166} \\
        \midrule
        \multicolumn{10}{l}{\textit{HANDAL: estimated depth}} \\
        \rowcolor{black!10}
        GPT-6-Astra & 0.032 & 100.00 & 0.012 & 0.787 & 0.861 & 0.674 & 19.10 & 0.914 & 0.106 \\
        SimFoundry & 0.134 & 63.64 & 0.026 & 0.723 & 0.624 & 0.572 & 17.16 & 0.915 & 0.143 \\
        RecGen & 0.080 & 69.09 & 0.020 & 0.785 & 0.695 & 0.624 & 17.38 & 0.914 & 0.137 \\
        ReconViaGen & --- & --- & 0.024 & 0.732 & 0.671 & 0.617 & 14.85 & 0.893 & 0.147 \\
        ShapeR & 0.090 & 70.91 & 0.030 & 0.658 & 0.581 & 0.539 & --- & --- & --- \\
        MV-SAM3D & 0.063 & \secondresult{80.00} & 0.017 & 0.766 & 0.754 & 0.647 & 15.98 & 0.904 & 0.124 \\
        Pixal3D & --- & --- & 0.014 & 0.761 & 0.828 & 0.686 & 17.84 & \secondresult{0.917} & 0.122 \\
        \modelname{} w/o agent (Ours) & \bestresult{0.032} & \bestresult{96.36} & \bestresult{0.008} & \secondresult{0.826} & \bestresult{0.936} & \bestresult{0.757} & \secondresult{19.42} & \bestresult{0.927} & \secondresult{0.097} \\
        \modelname{} (Ours) & \secondresult{0.033} & \bestresult{96.36} & \secondresult{0.009} & \bestresult{0.832} & \secondresult{0.923} & \secondresult{0.728} & \bestresult{20.07} & \secondresult{0.917} & \bestresult{0.096} \\
        \bottomrule
    \end{tabularx}
\end{table}

%% file: figures/bop_additional.tex
\begin{figure}[p]
    \centering
    \includegraphics[width=\linewidth,height=0.85\textheight,keepaspectratio]{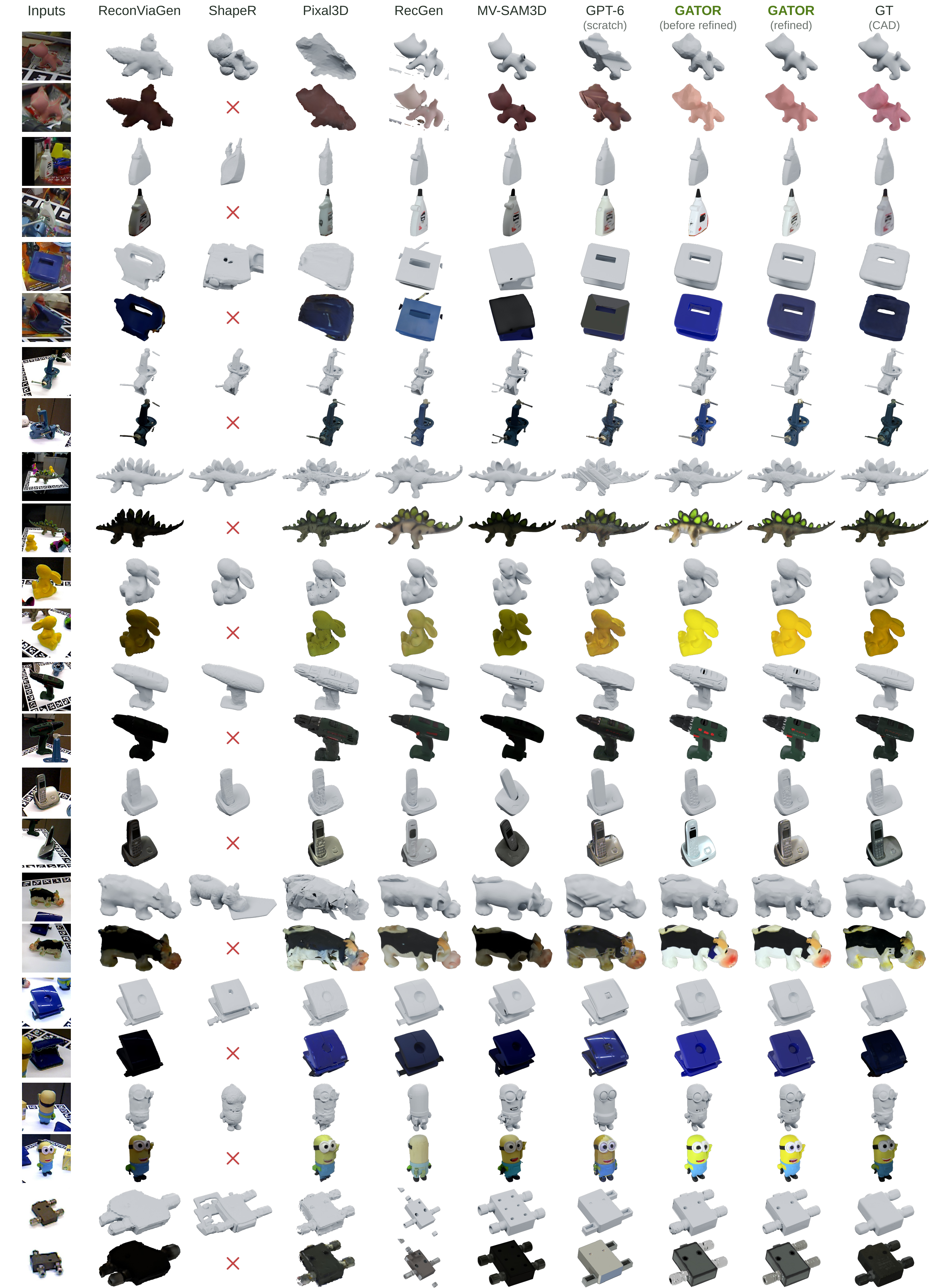}
    \caption{\textbf{Extended LM-O and HB comparisons.}
    Each object shows two of its four input views and each method's geometry (top) and appearance (bottom) from a shared camera.
    ShapeR can only output untextured object meshes (red crosses).}
    \label{fig:bop_additional}
\end{figure}

%% file: tables/bop_multiview_results.tex
\begin{table}[!t]
    \centering
    \footnotesize
    \setlength{\tabcolsep}{0.65pt}
    \caption{\textbf{Multiview tabletop reconstruction on LM-O and HB.}
    Unlike Table~\ref{tab:real_world_mv4_by_dataset}, each target is evaluated with two different view selections, recall counts every attempt, and \modelname{} is reported without agentic refinement, so its four-view scores differ between the two tables.
    \modelname{} leads every joint and geometry metric at every view count, and its distance errors decrease as views are added.
    Pixal3D predicts no pose placement and thus has no joint (ADD-SB) result.
    Bold marks the best score.}
    \label{tab:bop_multiview_results}
    \begin{tabularx}{\linewidth}{@{}c@{\hspace{6pt}}l *{9}{>{\centering\arraybackslash}X}@{}}
        \toprule
        \multirow{2}{*}{Views} & \multirow{2}{*}{Methods}
        & \multicolumn{2}{c}{Joint (ADD-SB)}
        & \multicolumn{4}{c}{Geometry}
        & \multicolumn{3}{c}{Appearance} \\
        \cmidrule(lr){3-4}\cmidrule(lr){5-8}\cmidrule(lr){9-11}
        & & Mean$\downarrow$ & @0.1$\uparrow$
        & CD$\downarrow$ & NC$\uparrow$ & F1$\uparrow$ & IoU$_{2\mathrm{D}}\!\uparrow$
        & PSNR$\uparrow$ & SSIM$\uparrow$ & LPIPS$\downarrow$ \\
        \midrule
        \multirow{3}{*}{2} & RecGen & 0.041 & \textbf{98.03} & 0.021 & 0.796 & 0.643 & 0.845 & 15.16 & \textbf{0.867} & 0.203 \\
          & Pixal3D & --- & --- & 0.027 & 0.757 & 0.648 & 0.840 & 15.32 & 0.865 & 0.209 \\
          & \modelname{} (Ours) & \textbf{0.015} & \textbf{98.03} & \textbf{0.011} & \textbf{0.865} & \textbf{0.863} & \textbf{0.897} & \textbf{15.75} & 0.866 & \textbf{0.183} \\
        \cmidrule(lr){1-11}
        \multirow{2}{*}{4} & Pixal3D & --- & --- & 0.021 & 0.799 & 0.745 & 0.875 & \textbf{16.29} & \textbf{0.872} & 0.190 \\
          & \modelname{} (Ours) & \textbf{0.014} & \textbf{98.03} & \textbf{0.010} & \textbf{0.874} & \textbf{0.891} & \textbf{0.902} & 15.86 & 0.866 & \textbf{0.179} \\
        \cmidrule(lr){1-11}
        \multirow{2}{*}{8} & Pixal3D & --- & --- & 0.020 & 0.813 & 0.774 & 0.886 & \textbf{16.74} & \textbf{0.875} & 0.185 \\
          & \modelname{} (Ours) & \textbf{0.013} & \textbf{98.03} & \textbf{0.010} & \textbf{0.880} & \textbf{0.901} & \textbf{0.900} & 15.85 & 0.867 & \textbf{0.174} \\
        \bottomrule
    \end{tabularx}
\end{table}

%% file: figures/real_scannetpp_additional.tex
\begin{figure}[p]
    \centering
    \includegraphics[width=\linewidth,height=0.85\textheight,keepaspectratio]{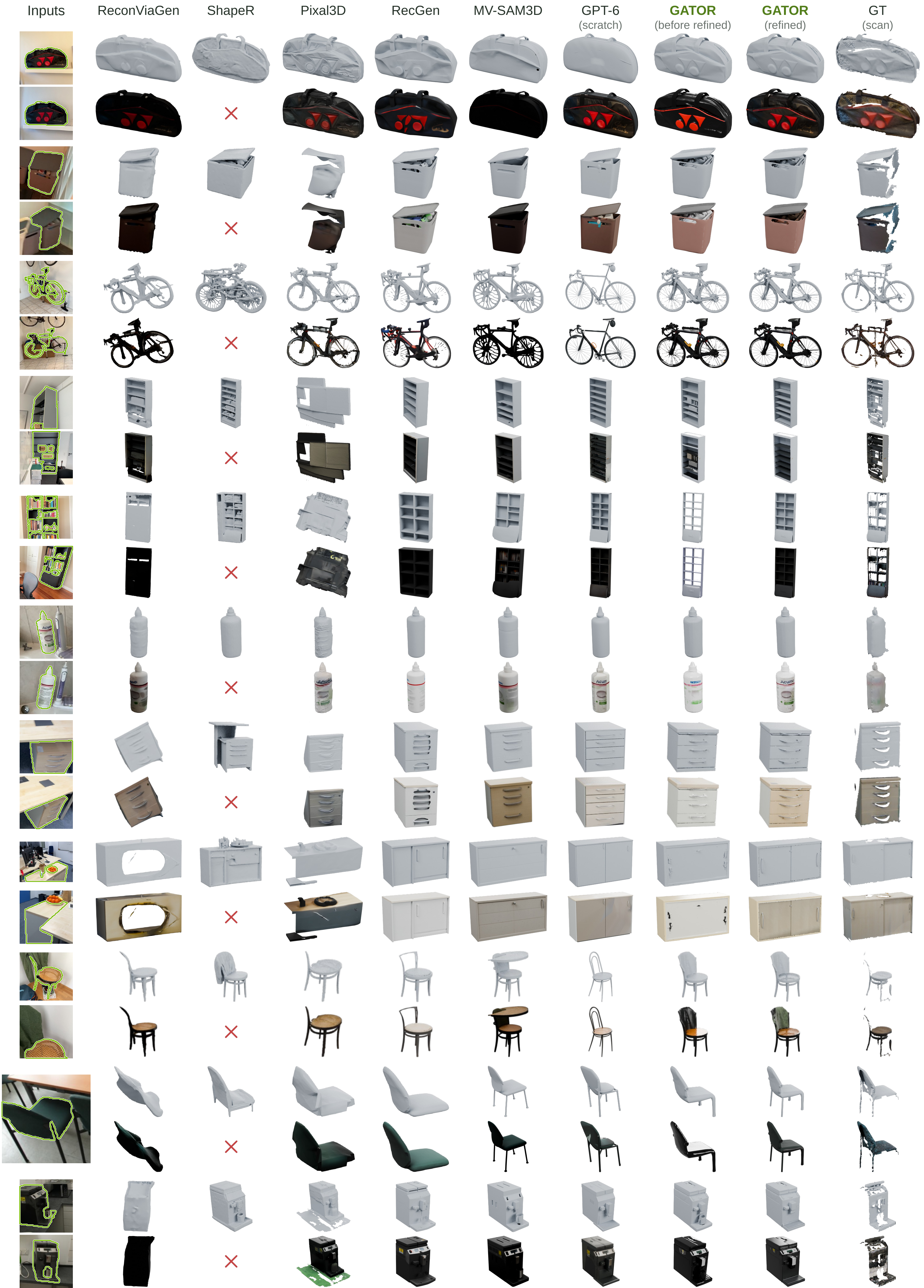}
    \caption{\textbf{Extended ScanNet++ comparisons.}
    Each object shows up to two input views with the target outlined in green and each method's geometry (top) and appearance (bottom) from a shared camera.
    ShapeR can only output untextured object meshes (red crosses). GT is the observed partial scan.}
    \label{fig:real_scannetpp_additional}
\end{figure}

%% file: figures/ablation.tex
\begin{figure}[t]
    \centering
    \includegraphics[width=\linewidth]{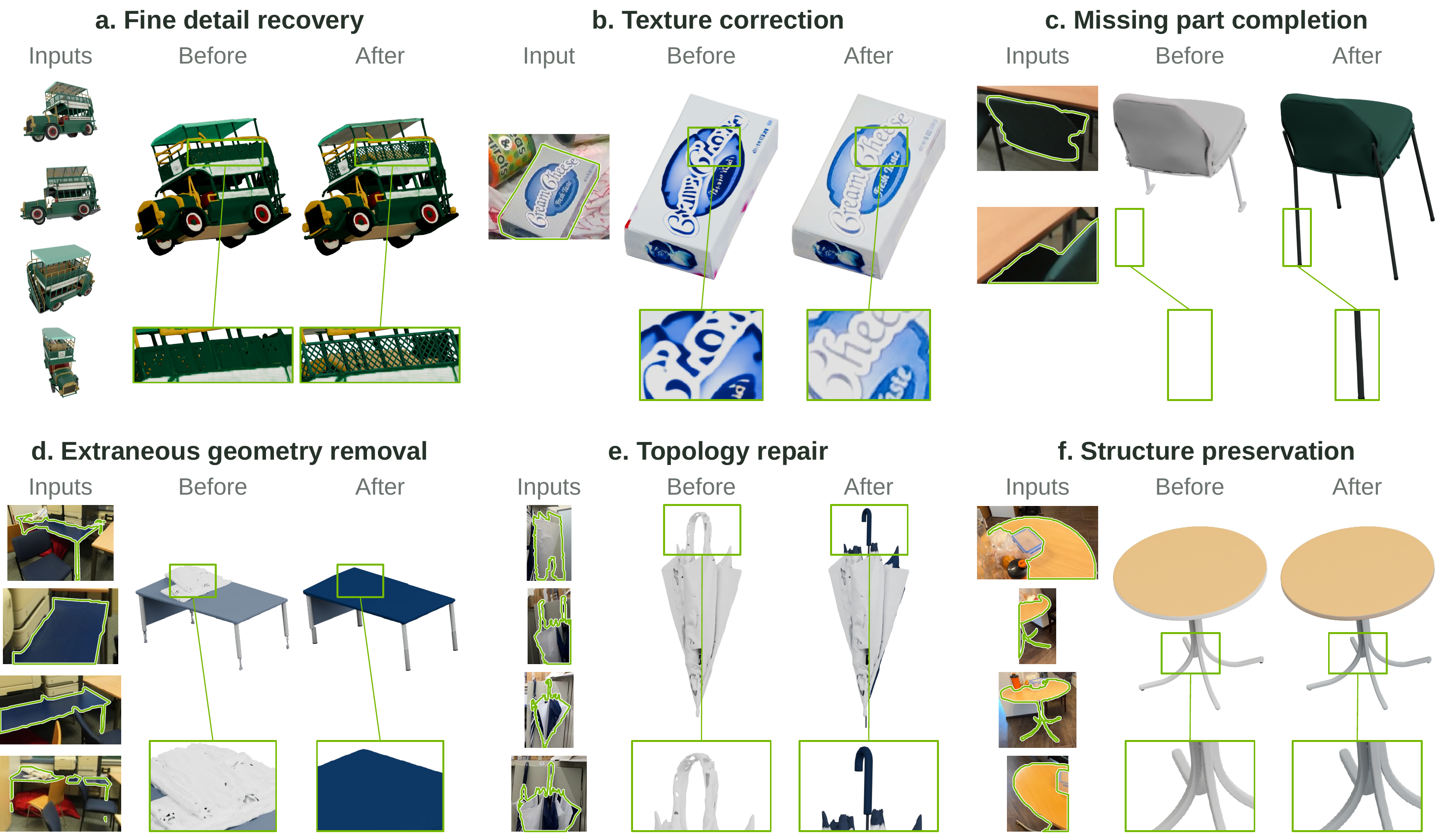}
    \caption{\textbf{Agentic reconstruction refinement.}
    Before-and-after examples illustrate six refinement behaviors.
    Up to four input views are shown, with real-world targets outlined in green.
    Boxes and matched zooms highlight reconstruction details.
    Each pair shares the same camera, diffuse shading, and display scale.}
    \label{fig:ablation_qualitative}
\end{figure}

%% file: figures/sim_readiness.tex
\begin{figure}[!htbp]
    \centering
    \includegraphics[width=\linewidth]{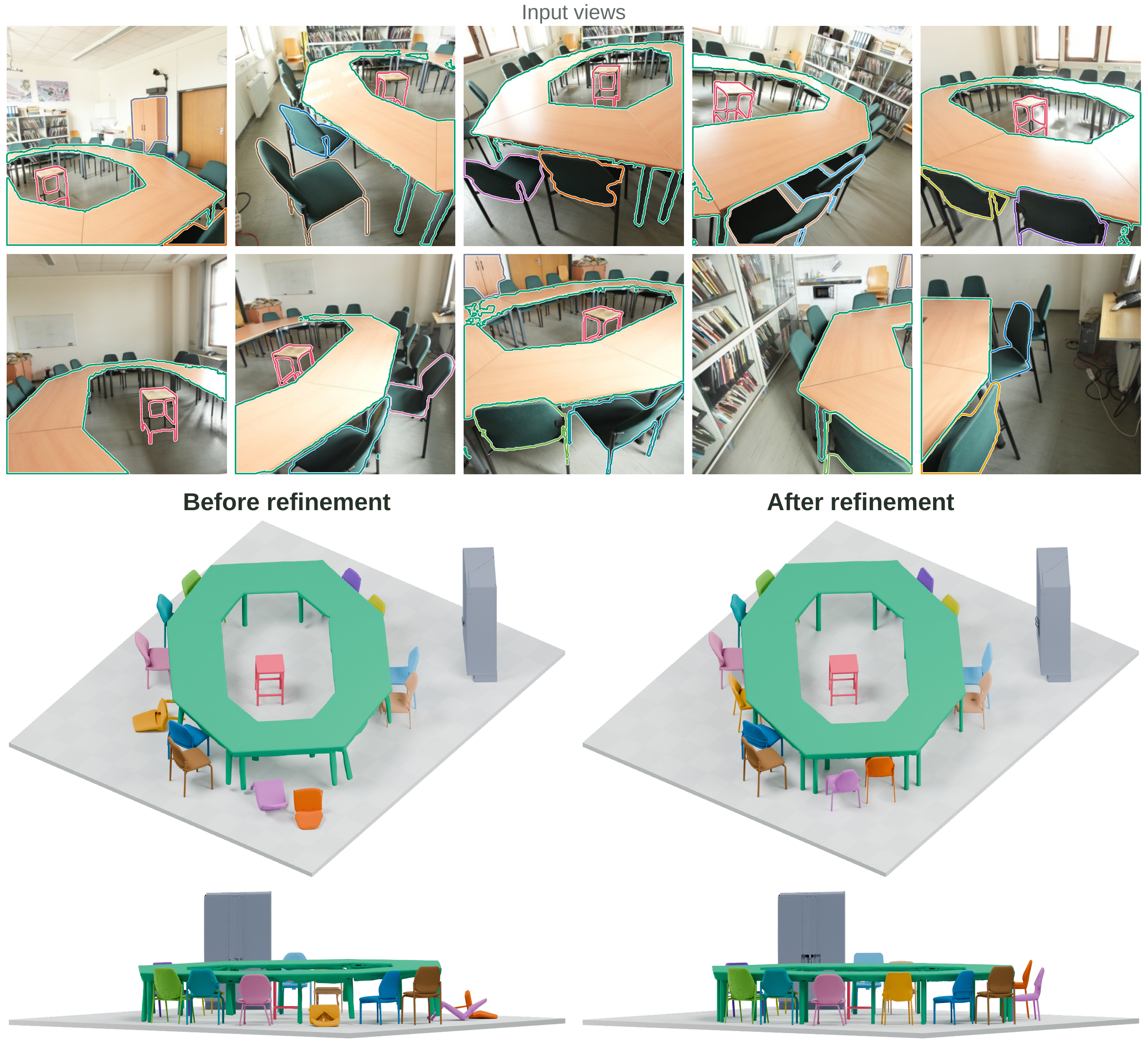}
    \caption{\textbf{Scene-level rigid-body simulation.}
    All ten input views (top two rows) and simulated resting states of a ScanNet++ meeting room, shown from overview (third row) and side (bottom) cameras.
    Three chairs that lean or topple before refinement become able to stand upright steadily after refinement.}
    \label{fig:sim_readiness}
\end{figure}